\documentclass{article} 
\usepackage{iclr2027_arxiv,times}

\usepackage{amsmath,amsfonts,bm}

\def\eqref#1{equation~\ref{#1}}

\def\1{\bm{1}}

\DeclareMathAlphabet{\mathsfit}{\encodingdefault}{\sfdefault}{m}{sl}
\SetMathAlphabet{\mathsfit}{bold}{\encodingdefault}{\sfdefault}{bx}{n}

\usepackage{hyperref}
\usepackage{url}
\usepackage{booktabs}
\usepackage{longtable}
\usepackage{multirow}
\usepackage{makecell}
\usepackage{array}
\usepackage{ragged2e}
\usepackage[table]{xcolor}
\usepackage{xltabular}
\usepackage{graphicx}
\usepackage{placeins}
\usepackage{capt-of}
\usepackage{float}
\hypersetup{
  pdftitle={
    GeoNIRS: Geometry-Aware Pretraining for Heterogeneous
    fNIRS Representation Learning
  },
  pdfauthor={
    Zongsheng Li, Shengyuan Zhu, Zimo Zhang,
    Hanyao Li, Shengming Wang, Quanying Liu
  },
  pdfkeywords={
    fNIRS, pretraining, representation learning,
    channel geometry, self-supervised learning
  }
}
\definecolor{tablecyan}{RGB}{229,247,247}
\definecolor{headergray}{RGB}{242,242,242}
\usepackage{xcolor}

\usepackage{colortbl}

\title{GeoNIRS: Geometry-Aware Pretraining for Heterogeneous fNIRS Representation Learning}

\author{
\begin{tabular}{c}
\textbf{Zongsheng Li}$^{1,2\dagger}$
\quad
\textbf{Shengyuan Zhu}$^{1\dagger}$
\quad
\textbf{Zimo Zhang}$^{1}$
\\
\textbf{Hanyao Li}$^{1}$
\quad
\textbf{Shengming Wang}$^{1}$
\quad
\textbf{Quanying Liu}$^{1,3,4\ast}$
\\[4pt]
\normalfont
$^{1}$Southern University of
Science and Technology, Shenzhen, China
\\
\normalfont
$^{2}$The Chinese University of Hong Kong, Shenzhen, China
\\
\normalfont
$^{3}$Omni-Intelligence, Shenzhen, China
\\
\normalfont
$^{4}$Shenzhen Loop Area Institute, Shenzhen, China
\\[3pt]
\normalfont
$^{\dagger}$Equal contribution
\\
\normalfont
$^{\ast}$Corresponding author:
\texttt{liuqy@sustech.edu.cn}
\end{tabular}
}

\iclrfinalcopy

\begin{document}

\maketitle






\begin{abstract}
Functional near-infrared spectroscopy (fNIRS) provides portable, low-cost, and spatially localized measurements of cortical hemodynamics, but substantial differences in channel count, probe layout, signal distribution, and task type hinder joint pretraining across datasets. We introduce GeoNIRS, a geometry-aware pretraining framework that uses the physical locations of source--detector pairs as a shared spatial reference across heterogeneous probe layouts. GeoNIRS compresses variable-size channel sets into a fixed-size latent representation and employs bidirectional HbO/HbR cross-reconstruction with contiguous temporal masking to learn transferable representations. GeoNIRS-base is pretrained on eight heterogeneous datasets, while GeoNIRS-large scales pretraining to 22 datasets. Neither the data nor labels of the downstream tasks are used during pretraining, enabling evaluation under a strict cross-task setting. To the best of our knowledge, GeoNIRS is the first fNIRS framework to explicitly model physical channel geometry during joint pretraining across heterogeneous channel configurations, and the first to provide both complete source code and downloadable pretrained checkpoints.

We evaluate GeoNIRS on eight held-out tasks spanning clinical and auditory assessment, pain and affective processing, semantic decoding, and motor function. GeoNIRS achieves the highest task-averaged balanced accuracy among the evaluated conventional machine-learning, fNIRS-specific, general time-series, and fNIRS pretraining baselines. Compared with random initialization, pretraining improves task-averaged balanced accuracy by 4.12 percentage points under linear probing and 3.30 points under full fine-tuning, with larger benefits when downstream labels are scarce. GeoNIRS also remains robust when input channels are missing. Geometry perturbations and pretraining ablations further demonstrate that both channel geometry and cross-chromophore reconstruction contribute to the learned representations. Overall, GeoNIRS provides an open and reproducible framework for transferable representation learning across heterogeneous fNIRS datasets. Our code and pretrained checkpoints are available at \url{https://github.com/GeoNIRS/GeoNIRS}
\end{abstract}

\section{Introduction}

Functional near-infrared spectroscopy (fNIRS) noninvasively measures changes in cortical oxygenated and deoxygenated hemoglobin (HbO/HbR), providing a hemodynamic proxy for neural activity \citep{ferrari2012brief,scholkmann2014review}. Compared with functional magnetic resonance imaging, fNIRS is relatively inexpensive, portable, and silent, supporting bedside monitoring, measurements during movement, and experiments in naturalistic settings. Its spatially localized cortical signals also complement the high-temporal-resolution electrophysiological activity recorded by EEG \citep{pinti2020present,yucel2021best}. These properties have enabled applications across cognitive and affective neuroscience, brain--computer interfaces, motor and neurorehabilitation research, and the assessment of neurological and psychiatric conditions \citep{obrig2014nirs,ehlis2014application,naseer2015fnirs,mihara2016review}. In the closely related EEG domain, neural foundation models and their benchmarking ecosystems are considerably more mature. OmniEEG-Bench and Meta FAIR's NeuralBench-EEG systematically evaluate pretrained models across multiple datasets, tasks, and standardized adaptation protocols \citep{lu2026omnieeg,banville2026neuralbench}. BENDR provided early evidence for large-scale self-supervised EEG pretraining, while LaBraM, EEGPT, BrainOmni, and REVE further demonstrated that aggregating unlabeled recordings across datasets, tasks, and acquisition systems can produce transferable neural representations \citep{kostas2021bendr,jiang2024large,wang2024eegpt,xiao2025brainomni,elouahidi2025reve}. Such multi-dataset pretraining alleviates the limited scale of individual datasets while exposing models to variations in populations, paradigms, and acquisition systems, thereby facilitating transfer to unseen tasks and label-scarce settings. Extending this paradigm to heterogeneous fNIRS datasets is therefore an important step from dataset-specific decoding toward general-purpose neural representation learning.

Conventional fNIRS decoding remains largely dataset-specific and supervised, ranging from handcrafted HbO/HbR features with LDA or SVM to fNIRS-specific deep networks \citep{naseer2016lda,naseer2016classification,wang2022transformer,wang2023fnirsnet,wang2026fnirs4d}. These methods are typically trained separately under fixed probe layouts and label spaces, preventing joint learning from heterogeneous unlabeled data. Extending multi-dataset pretraining to fNIRS presents two key challenges. First, channel semantics vary across probe layouts: the same channel index may correspond to different cortical locations, while the same region may be covered by different channels. Variable-channel support alone therefore cannot establish cross-layout spatial correspondence. Second, designing a non-trivial pretraining objective is difficult. Slowly varying hemodynamic signals make within-signal masked reconstruction susceptible to temporal interpolation \citep{scholkmann2014review,zhou2026xmae}. Self-reconstruction within each chromophore may primarily capture modality-specific signal structure. In contrast, cross-chromophore reconstruction encourages the model to leverage the shared physiological information between HbO and HbR, promoting more transferable representations \citep{cui2010functional}.
As shown in Table~\ref{tab:pretraining_framework_comparison}, existing methods address only parts of these challenges. Early approaches use a single dataset or train separately across datasets \citep{liu2021unsupervised,wang2021taming,wang2024empower}. Jung et al. jointly pretrain across varying channel counts but do not model physical channel geometry \citep{jung2025fnirs}; EFRM uses only one fNIRS dataset \citep{jung2025efrm}; and Subedi et al. assume a shared probe configuration and require supervised adaptation to new layouts \citep{subedi2026interpretable}. ASDGPrompt and DSSER likewise do not perform joint multi-dataset pretraining \citep{cai2026asdgprompt,wang2026dsser}. Thus, existing work has yet to jointly address cross-layout spatial unification and physiologically grounded pretraining, while reproducibility remains limited by the scarcity of complete source code and downloadable pretrained checkpoints.

The design of our work is inspired by recent advances in EEG foundation models and physiological signal pretraining. First, BrainOmni shows that incorporating the physical properties of sensors into representation learning and compressing a variable number of sensor inputs into a fixed-size latent representation can mitigate structural differences across acquisition systems \citep{xiao2025brainomni}. Second, xMAE performs masked cross-modal reconstruction between synchronously recorded ECG and PPG signals, demonstrating that predicting masked signals from related physiological observations can reduce reliance on local temporal smoothness \citep{zhou2026xmae}. However, each fNIRS channel is defined by a source--detector pair, while HbO and HbR are coupled but non-equivalent hemodynamic observations within the same modality \citep{cui2010functional}. These ideas must therefore be redesigned for fNIRS to unify heterogeneous probe layouts and learn the hemodynamic structure shared between HbO and HbR.

Building on these ideas, we propose GeoNIRS, a geometry-aware pretraining framework for heterogeneous fNIRS data. GeoNIRS explicitly encodes source--detector geometry and compresses variable-size channel sets from heterogeneous probe layouts into a fixed-size latent representation. It is pretrained via bidirectional HbO/HbR cross-reconstruction with contiguous temporal masking, encouraging the model to capture shared hemodynamic structure while reducing reliance on local temporal interpolation. GeoNIRS-base is pretrained on 383.30 hours of recordings across eight datasets, while GeoNIRS-large scales pretraining to 827.04 hours across
22 datasets. To the best of our knowledge, GeoNIRS is the first fNIRS framework to explicitly incorporate physical channel geometry for joint pretraining across heterogeneous probe layouts, and the first to release both complete source code and downloadable pretrained checkpoints. Across eight held-out downstream tasks, GeoNIRS achieves the strongest task-averaged performance among the evaluated methods. Compared with random initialization, pretraining improves average balanced accuracy by 4.12 percentage points under linear probing and 3.30 percentage points under full fine-tuning. Further experiments show larger pretraining gains in low-label regimes, robustness to missing channels, and task-averaged benefits from channel geometry and cross-chromophore reconstruction across all three adaptation protocols.

\begin{figure}[t]
    \centering
    \includegraphics[width=1.1\linewidth]{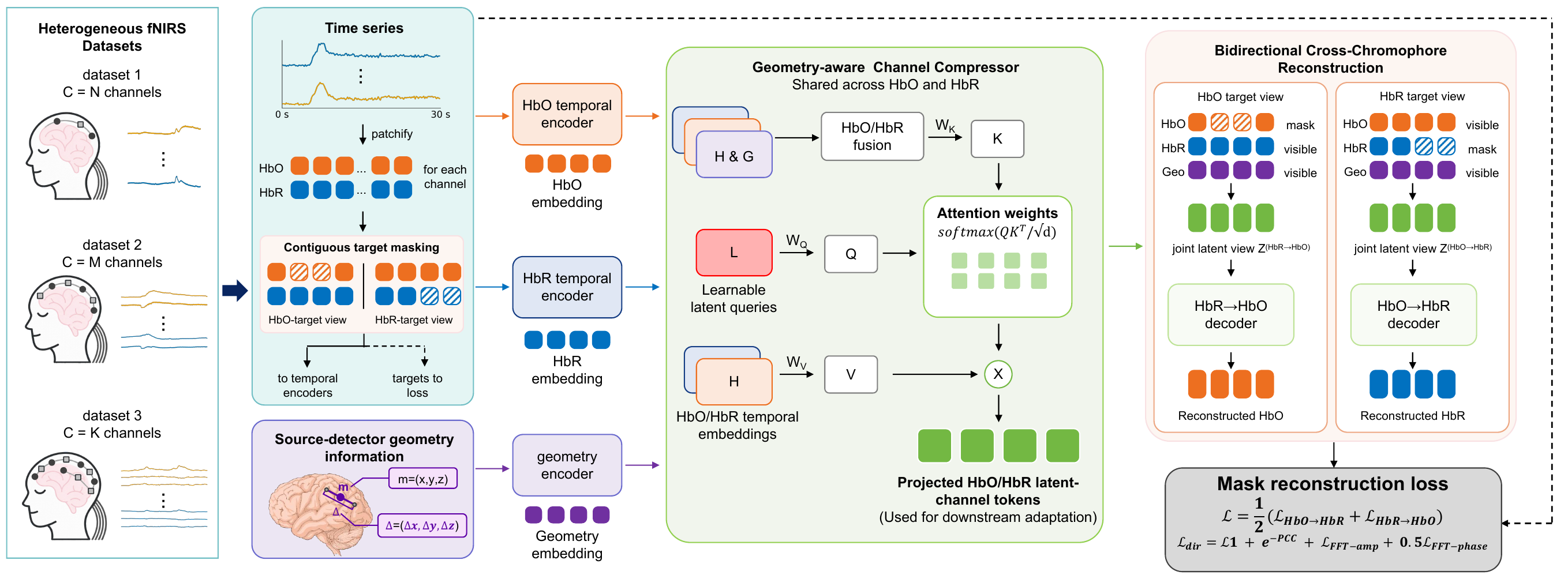}
    \caption{
   Overview of the GeoNIRS framework. HbO/HbR signals from datasets with varying channel counts and probe layouts are divided into temporal patches and encoded alongside source--detector geometry. The geometry-aware channel compressor uses learnable latent queries and shared geometry-aware channel attention to map variable-size physical channel sets into a fixed number of latent-channel tokens. Pretraining employs bidirectional cross-chromophore reconstruction: contiguous patches of the target chromophore are masked and reconstructed using its visible context together with the complete complementary chromophore. The reconstruction loss is computed only at masked and valid target positions. After pretraining, the encoder and channel compressor are used for downstream adaptation.
    }
    \label{fig:framework}
\end{figure}

\section{Method}
\label{sec:method}

\subsection{Overview}
\label{sec:method_overview}


As shown in Figure~\ref{fig:framework}, GeoNIRS processes paired HbO/HbR
signals and source--detector geometry from datasets with varying channel
counts and probe layouts. Separate temporal encoders represent the two
chromophores, while a geometry-aware channel compressor maps each
dataset-specific physical-channel set into a fixed number of latent-channel
tokens. Pretraining masks contiguous patches of one chromophore and reconstructs
them using its visible context together with the complete complementary
chromophore, with the two reconstruction directions optimized jointly. After
pretraining, the reconstruction decoders are discarded and the fixed-length
representations are adapted through linear probing, LoRA, or full fine-tuning.

\subsection{Temporal and Geometric Representations}

For a recording from dataset \(d\), we denote the paired chromophore signals as \(X_d^{\mathrm{O}}, X_d^{\mathrm{R}} \in \mathbb{R}^{C_d \times T}\), where \(C_d\) is the dataset-specific number of physical channels and may vary across datasets. Along the temporal dimension, each channel is divided into \(P\) non-overlapping patches, each containing \(S\) samples. Every patch is projected into a \(D\)-dimensional token and augmented with a learnable temporal positional embedding. Separate HbO and HbR temporal encoders then model the token sequence within each physical channel, producing \(H^{\mathrm{O}}, H^{\mathrm{R}} \in \mathbb{R}^{C_d \times P \times D}\). Because each temporal encoder is applied independently to all channels with weights shared across channels, the model does not require a fixed \(C_d\). Samples and channels rejected during preprocessing quality control, together with patches containing insufficient valid samples, are excluded using validity masks.

\subsection{Geometry-Aware Channel Compression}
\label{sec:channel_compression}

To transform variable-size physical channel sets into fixed-length representations, we introduce \(M\) learnable latent queries, denoted by \(L\in\mathbb{R}^{M\times D}\). For the \(p\)-th temporal patch, the paired HbO and HbR representations are first fused as
\[
\bar{H}_{p}
=
\frac{1}{2}
\left(
H_{:,p}^{\mathrm{O}}
+
H_{:,p}^{\mathrm{R}}
\right).
\]
The latent queries are projected into queries, while the fused temporal representations and channel geometry jointly form the keys:
\[
Q = W_Q\operatorname{LN}(L),
\qquad
K_p
=
W_K\operatorname{LN}
\left(
\bar{H}_{p}+G
\right),
\]
where \(G\in\mathbb{R}^{C_d\times D}\) denotes the channel representations produced by the geometry encoder. The attention weights from the fixed latent channels to the current physical channel set are then computed as
\[
A_p
=
\operatorname{softmax}
\left(
\frac{QK_p^\top}{\sqrt{d_h}}
\right)
\in\mathbb{R}^{M\times C_d},
\]
where \(d_h\) is the attention-head dimension. Invalid channels are masked before the softmax operation.

HbO and HbR share the same attention weights but retain separate value representations:
\[
V_p^{s}
=
W_V\operatorname{LN}
\left(
H_{:,p}^{s}
\right),
\qquad
Z_p^{s}
=
A_pV_p^{s},
\quad
s\in\{\mathrm{O},\mathrm{R}\}.
\]
This design gives both chromophores a shared spatial mapping determined jointly by their paired signals and physical geometry, while preserving their chromophore-specific content. After an output projection and a feed-forward block, the representations across all temporal patches form
$Z^{\mathrm O}, Z^{\mathrm R} \in \mathbb{R}^{M\times P\times F}$, where $F$ denotes the output feature dimension.

\subsection{Bidirectional Cross-Chromophore Reconstruction}
\label{sec:cross_reconstruction}

GeoNIRS employs asymmetric bidirectional cross-chromophore reconstruction. For each recording, a contiguous span of valid temporal patches, denoted by \(\Omega\), is randomly selected as the reconstruction target. In the HbO\(\rightarrow\)HbR direction, HbO remains fully visible while the HbR patches within \(\Omega\) are masked. The two streams jointly determine the geometry-aware channel attention, while the complete HbO representation provides the values to be aggregated. The reverse direction uses the same target span, keeps HbR fully visible, and masks the corresponding HbO patches. The compact representations for the two directions are
\[
Z^{\mathrm{O}\rightarrow\mathrm{R}}
=
\Pi^{\mathrm{O}}
\left[
A\!\left(
H^{\mathrm{O}},
\widetilde{H}^{\mathrm{R}},
G
\right)
V\!\left(H^{\mathrm{O}}\right)
\right],
\]
\[
Z^{\mathrm{R}\rightarrow\mathrm{O}}
=
\Pi^{\mathrm{R}}
\left[
A\!\left(
\widetilde{H}^{\mathrm{O}},
H^{\mathrm{R}},
G
\right)
V\!\left(H^{\mathrm{R}}\right)
\right],
\]
where \(\widetilde{H}\) denotes a temporally masked target representation, \(A(\cdot)\) is the geometry-aware channel attention defined above, and $\Pi_s$ is a chromophore-specific residual projection that maps the feature dimension from $D$ to $F$ without mixing latent channels or
temporal patches. The visible context of the target chromophore therefore contributes to channel-attention estimation, while the complete complementary chromophore supplies the reconstruction content. Contiguous masking further reduces reliance on local temporal interpolation.

The cross-chromophore decoders map the fixed latent-channel representations back to the \(C_d\) physical channels of each dataset. For physical channel \(c\) and temporal patch \(p\), the decoder query combines the channel geometry, a temporal positional embedding, and a learnable bias:
\[
q_{c,p}^{\mathrm{dec}}
=
G_c
+
e_p^{\mathrm{dec}}
+
b^{\mathrm{dec}}.
\]
These queries retrieve reconstruction features from the corresponding latent tokens through cross-attention. A channel-wise temporal decoder and reconstruction head then produce the original signal patches:
\[
\widehat{X}^{\mathrm{R}}
=
\mathcal{D}_{\mathrm{O}\rightarrow\mathrm{R}}
\left(
Z^{\mathrm{O}\rightarrow\mathrm{R}},G
\right),
\qquad
\widehat{X}^{\mathrm{O}}
=
\mathcal{D}_{\mathrm{R}\rightarrow\mathrm{O}}
\left(
Z^{\mathrm{R}\rightarrow\mathrm{O}},G
\right).
\]
Because the physical-channel queries are constructed from the corresponding geometry, the same decoder can generate channel-specific reconstructions for datasets with different channel counts and probe layouts.

Each reconstruction direction is optimized using complementary time- and frequency-domain objectives:
\[
\mathcal{L}_{\mathrm{dir}}
=
\mathcal{L}_{1}
+
\mathcal{L}_{\mathrm{PCC}}
+
\mathcal{L}_{\mathrm{amp}}
+
\frac{1}{2}\mathcal{L}_{\mathrm{phase}},
\qquad
\mathcal{L}_{\mathrm{PCC}}
=
\exp\!\left[
-\operatorname{PCC}
\left(
\widehat{X},X
\right)
\right].
\]
Here, \(\mathcal{L}_{1}\) measures the time-domain absolute error, while \(\mathcal{L}_{\mathrm{amp}}\) and \(\mathcal{L}_{\mathrm{phase}}\) measure the Fourier-amplitude error and circular phase discrepancy, respectively. All terms are computed only over target samples that are both masked and valid after quality control. The final pretraining objective averages the two reconstruction directions:
\[
\mathcal{L}_{\mathrm{pre}}
=
\frac{1}{2}
\left(
\mathcal{L}_{\mathrm{O}\rightarrow\mathrm{R}}
+
\mathcal{L}_{\mathrm{R}\rightarrow\mathrm{O}}
\right).
\]

\subsection{Downstream Adaptation}

During downstream adaptation, temporal masking is disabled and the reconstruction decoders are discarded. The complete HbO/HbR signals and their channel geometry are processed by the pretrained temporal encoders, geometry encoder, and channel compressor, producing fixed-length representations \(Z^{\mathrm{O}}, Z^{\mathrm{R}} \in \mathbb{R}^{M \times P \times F}\). The two chromophore representations are then concatenated along the feature dimension:
\[
Z^{\mathrm{joint}}
=
\left[
Z^{\mathrm{O}}
\mathbin{\|}
Z^{\mathrm{R}}
\right]
\in
\mathbb{R}^{M \times P \times 2F},
\]
and passed to a task-specific prediction head. For linear probing, the pretrained backbone is frozen and only the prediction head is trained. LoRA updates the low-rank adaptation parameters and prediction head, whereas full fine-tuning jointly updates the entire pretrained backbone and downstream head.
\begin{table*}[t]
\centering
\caption{
Cross-task fNIRS decoding performance measured by balanced accuracy (\%).
Results are averaged over two random seeds, with subject-level five-fold
cross-validation performed for each seed. The best and second-best results
in each column are highlighted in bold and underline, respectively.
}
\label{tab:benchmark}

\setlength{\tabcolsep}{1.5pt}
\renewcommand{\arraystretch}{1.08}
\footnotesize

\begin{tabular*}{\linewidth}{
@{\extracolsep{\fill}}
l*{9}{c}
@{}
}
\toprule

\textbf{Method}
& \textbf{ICH}
& \makecell{\textbf{CI}\\\textbf{Status}}
& \textbf{Tinnitus}
& \makecell{\textbf{Pain}\\\textbf{Intensity}}
& \textbf{Emotion}
& \textbf{Semantic}
& \makecell{\textbf{Motor}\\\textbf{Recovery}}
& \makecell{\textbf{Action}\\\textbf{Mode}}
& \textbf{Average}
\\

\midrule

RBF-SVM~\citeyearpar{naseer2016classification}
& 74.83
& 64.14
& 55.65
& 52.84
& 56.01
& 50.36
& 57.82
& 59.65
& 58.91
\\

Stat-LDA~\citeyearpar{naseer2016lda}
& 51.55
& 53.09
& 51.53
& 54.66
& \textbf{59.38}
& 50.61
& 62.40
& 65.94
& 56.14
\\

\midrule

1D-CNN~\citeyearpar{trakoolwilaiwan2018cnn}
& 66.42
& 57.67
& 56.88
& 50.60
& 58.09
& 50.63
& 55.63
& 56.14
& 56.51
\\

EEGNet~\citeyearpar{lawhern2018eegnet}
& 56.35
& 49.68
& 51.12
& 49.92
& 52.68
& 49.85
& 59.74
& 65.00
& 54.29
\\

fNIRS-T~\citeyearpar{wang2022transformer}
& 72.25
& 64.19
& 52.05
& 52.90
& 55.63
& 50.61
& \underline{64.41}
& 70.87
& 60.36
\\

fNIRSNet~\citeyearpar{wang2023fnirsnet}
& 52.05
& 48.69
& 50.00
& 50.00
& 49.93
& 50.00
& 55.37
& 50.00
& 50.75
\\

fNIRS4D~\citeyearpar{wang2026fnirs4d}
& 75.71
& 51.40
& 49.98
& 51.79
& 55.33
& 50.19
& 57.82
& 69.11
& 57.67
\\

\midrule

MOMENT-LP~\citeyearpar{goswami2024moment}
& 73.93
& 60.48
& 54.29
& 54.43
& 55.69
& 50.17
& 60.83
& 70.12
& 59.99
\\

MOMENT-LoRA~\citeyearpar{goswami2024moment}
& 72.16
& 62.10
& 53.47
& 53.95
& 55.03
& 50.20
& 63.28
& \textbf{71.12}
& 60.17
\\

\midrule

Jung-MAE-LP~\citeyearpar{jung2025fnirs}
& 72.21
& 65.45
& 51.76
& \underline{55.22}
& 58.14
& \textbf{51.48}
& 62.24
& 68.98
& 60.68
\\

Jung-MAE-LoRA~\citeyearpar{jung2025fnirs}
& 77.52
& 64.12
& 54.85
& 54.73
& 57.39
& 50.72
& 64.19
& 70.31
& 61.73
\\

\midrule

GeoNIRS-LP (ours)
& 77.82
& 68.06
& \underline{58.70}
& \textbf{56.19}
& 58.23
& 50.93
& 64.27
& 70.29
& 63.06
\\

GeoNIRS-LoRA (ours)
& \underline{80.10}
& \textbf{69.53}
& \textbf{58.72}
& 54.36
& \underline{59.15}
& \underline{51.22}
& 63.92
& 70.34
& \underline{63.42}
\\

GeoNIRS-FT (ours)
& \textbf{80.86}
& \underline{69.19}
& 57.70
& 54.70
& 59.13
& 51.03
& \textbf{64.75}
& \underline{70.99}
& \textbf{63.54}
\\
\bottomrule
\end{tabular*}

\vspace{3pt}

\begin{minipage}{0.98\textwidth}
\footnotesize
\textit{Note:}
ICH denotes intracerebral hemorrhage, and CI denotes cochlear implant. MOMENT-LP and MOMENT-LoRA both use the MOMENT-1-base backbone. Jung-MAE denotes our Jung-inspired MAE implementation. Average performance is computed equally across all eight tasks.
\end{minipage}

\end{table*}
\section{Experiments}
We conduct experiments to answer six questions:
(1) whether GeoNIRS outperforms existing methods on downstream fNIRS tasks
that are unseen during pretraining;
(2) whether its downstream gains arise from pretraining rather than the model
architecture alone;
(3) whether pretraining improves transfer when downstream labels are scarce;
(4) whether geometry-aware representations remain robust when input channels
are missing;
(5) how channel geometry and cross-chromophore reconstruction contribute to
the learned representations; and (6) how performance changes when scaling model capacity and pretraining data.
We first describe the experimental setup and then present cross-task
benchmarking, random-initialization controls, label-efficient adaptation,
channel-loss robustness, and component ablations.

\subsection{Experimental Setup}
\paragraph{Pretraining data.}
Our dataset inventory comprises 30 fNIRS datasets, of which 22 are used for pretraining and eight are held out exclusively for downstream evaluation. GeoNIRS-base is pretrained on eight datasets spanning cognitive and executive functions, social interaction, naturalistic stimulation, and functional localization. GeoNIRS-large extends this corpus with 14 additional datasets, resulting in a total of 22 pretraining datasets. Dataset-level descriptions, statistics, and references are provided in Supplementary Table~1.

The complete 22-dataset pretraining corpus contains 3,578 recordings totaling 827.04 hours and yields 93,726 30-s windows after preprocessing. All windows are resampled to 5 Hz. The datasets vary substantially in participant population, channel count, probe layout, acquisition system, and experimental paradigm. None of the eight downstream datasets is included in either pretraining corpus. Unless otherwise specified, GeoNIRS refers to GeoNIRS-base; results for GeoNIRS-large are reported in Appendix~\ref{app:model_scaling}. 

\paragraph{Downstream tasks.}

We evaluate GeoNIRS on eight held-out binary classification tasks spanning
clinical and auditory assessment, pain and affective processing, semantic
decoding, and motor function. All experiments follow a cross-subject setting:
each participant is assigned exclusively to the training, validation, or test
set, and all compared methods use identical subject assignments. Input-window
duration is configured separately for each task according to its experimental
paradigm. Complete task definitions, references, and dataset statistics are
provided in Appendix~\ref{app:datasets}.

\paragraph{Baselines.}

We compare GeoNIRS against four groups of baselines.
First, conventional machine-learning approaches include an RBF-SVM
\citep{naseer2016classification} and statistical-feature LDA
\citep{naseer2016lda}.
Second, supervised neural baselines include 1D-CNN
\citep{trakoolwilaiwan2018cnn}, EEGNet
\citep{lawhern2018eegnet}, fNIRS-T
\citep{wang2022transformer}, fNIRSNet
\citep{wang2023fnirsnet}, and fNIRS4D
\citep{wang2026fnirs4d}.
Third, MOMENT-1-base
\citep{goswami2024moment} serves as a general-purpose pretrained
time-series baseline and is evaluated using linear probing and LoRA.
Fourth, we evaluate a paper-guided implementation of the
multi-dataset fNIRS pretraining framework of \citet{jung2025fnirs}, denoted as
Jung-inspired MAE. To accommodate heterogeneous channel counts, Jung-inspired
MAE uses channel-independent encoding with shared parameters but does not model
physical channel geometry. All methods use the same task-specific input windows
and cross-subject splits. Implementation details are provided in
Appendix~\ref{app:implementation_details}, while the capabilities and availability of
existing fNIRS pretraining methods are summarized in
Table~\ref{tab:pretraining_framework_comparison}.

\paragraph{Adaptation protocols.}

We evaluate GeoNIRS under three downstream adaptation protocols.
For linear probing (LP), the pretrained backbone is frozen and only a
task-specific classification head is optimized.
For low-rank adaptation (LoRA), low-rank trainable matrices are introduced
into the encoder, while the remaining pretrained parameters
are kept frozen.
For full fine-tuning (FT), the entire backbone and classification head are
jointly optimized.
All three protocols use identical subject splits within each downstream
task.
The label-efficiency and channel-loss
experiments use full fine-tuning.
MOMENT-LP and MOMENT-LoRA both use the publicly released MOMENT-1-base
backbone.

\paragraph{Evaluation protocol.}

We perform subject-level five-fold cross-validation, such that subjects in
the test fold are never observed during downstream training or model
selection.
Within each fold, the remaining subjects are randomly divided into training and
validation sets.
Each experiment is repeated with two random seeds,
810 and 811, and the reported results are averaged across folds and
seeds.
Balanced accuracy is used as the primary evaluation metric because several
datasets exhibit class imbalance.
AUROC and macro-F1 are additionally reported in the pretraining-control
experiments and in the supplementary results.
Task-averaged results assign equal weight to each of the eight downstream
tasks, irrespective of their number of participants or windows.

Complete implementation details are provided in
Appendix~\ref{app:implementation_details}.

\subsection{Cross-Task Downstream Performance}

Table~\ref{tab:benchmark} reports balanced accuracy across the eight held-out downstream tasks. The three GeoNIRS adaptation variants achieve the highest overall averages among all evaluated methods: 63.54 for full fine-tuning, 63.42 for LoRA, and 63.06 for linear probing, compared with 61.73 for the strongest non-GeoNIRS baseline. Notably, linear probing already surpasses all baselines while keeping the pretrained backbone frozen, indicating that GeoNIRS learns representations that transfer directly across datasets and tasks. LoRA performs within 0.12 percentage points of full fine-tuning, showing that competitive transfer can also be achieved through parameter-efficient adaptation. The largest margins over the strongest non-GeoNIRS results occur on CI Status and ICH, where GeoNIRS improves balanced accuracy by 4.08 and 3.34 percentage points, respectively. GeoNIRS also ranks first on Tinnitus, Pain Intensity, and Motor Recovery, and second on Emotion, Semantic, and Action Mode. These results demonstrate broad cross-dataset transfer rather than an improvement driven by a single downstream dataset.

\subsection{Effect of Pretraining}
\label{sec:effect_of_pretraining}
Table~\ref{tab:pretrain_effect} isolates the contribution of pretraining by comparing GeoNIRS with an otherwise identical randomly initialized counterpart. Under linear probing, pretraining improves average balanced accuracy by 4.12 percentage points and AUROC by 5.48 points, with positive gains across all eight tasks. Under full fine-tuning, the corresponding average gains are 3.30 and 3.96 points, respectively.

The larger gain under linear probing is expected because the frozen backbone more directly reflects representation quality, whereas full fine-tuning can partially compensate for random initialization using downstream labels. Because the compared models share the same architecture, data splits, and adaptation protocol, the observed gains can be attributed to pretraining rather than model capacity.

\begin{table}[t]
\centering
\caption{
Effect of fNIRS pretraining on downstream transfer performance across
eight tasks. Results are reported as balanced accuracy (\%).
$\Delta$ denotes the absolute improvement brought by pretraining,
calculated as Pretrained minus Random Init in percentage points.
}
\label{tab:pretrain_effect}

\setlength{\tabcolsep}{1.0pt}
\renewcommand{\arraystretch}{1.08}
\footnotesize

\begin{tabular*}{\linewidth}{
@{\extracolsep{\fill}}
l*{8}{c}
@{}
}
\toprule

&
\multicolumn{4}{c}{\textbf{Linear Probing}}
&
\multicolumn{4}{c}{\textbf{Full Fine-Tuning}}
\\

\cmidrule(lr){2-5}
\cmidrule(lr){6-9}

\makecell[l]{\textbf{Downstream}\\\textbf{Task}}
&
\makecell{\textbf{Random}\\\textbf{Init}}
&
\textbf{Pretrained}
&
\makecell{\boldmath$\Delta_{\mathrm{BAcc}}$\\\textbf{(pp)}}
&
\makecell{\boldmath$\Delta_{\mathrm{AUROC}}$\\\textbf{(pp)}}
&
\makecell{\textbf{Random}\\\textbf{Init}}
&
\textbf{Pretrained}
&
\makecell{\boldmath$\Delta_{\mathrm{BAcc}}$\\\textbf{(pp)}}
&
\makecell{\boldmath$\Delta_{\mathrm{AUROC}}$\\\textbf{(pp)}}
\\

\midrule

ICH
& 71.68
& \textbf{77.82}
& +6.14
& +6.72
& 75.18
& \textbf{80.86}
& +5.68
& +5.13
\\

\makecell[l]{CI Status}
& 64.32
& \textbf{68.06}
& +3.74
& +5.03
& 66.50
& \textbf{69.19}
& +2.69
& +2.46
\\

Tinnitus
& 53.31
& \textbf{58.70}
& +5.39
& +10.17
& 51.85
& \textbf{57.70}
& +5.85
& +8.96
\\

\makecell[l]{Pain Intensity}
& 52.33
& \textbf{56.19}
& +3.86
& +3.34
& 52.40
& \textbf{54.70}
& +2.30
& +0.07
\\

Emotion
& 54.68
& \textbf{58.23}
& +3.55
& +4.43
& 54.66
& \textbf{59.13}
& +4.47
& +6.63
\\

Semantic
& 49.55
& \textbf{50.93}
& +1.38
& +1.43
& \textbf{51.52}
& 51.03
& $-0.49$
& $-1.15$
\\

\makecell[l]{Motor Recovery}
& 60.33
& \textbf{64.27}
& +3.94
& +5.56
& 63.14
& \textbf{64.75}
& +1.61
& +3.52
\\

\makecell[l]{Action Mode}
& 65.30
& \textbf{70.29}
& +4.99
& +7.14
& 66.72
& \textbf{70.99}
& +4.27
& +6.09
\\

\midrule

\textbf{Average}
& 58.94
& \textbf{63.06}
& +4.12
& +5.48
& 60.25
& \textbf{63.54}
& +3.30
& +3.96
\\

\bottomrule
\end{tabular*}

\vspace{2pt}

\begin{minipage}{\linewidth}
\footnotesize
\end{minipage}

\end{table}

\FloatBarrier

\subsection{Label-Efficient Adaptation}
\label{sec:label_efficiency}

\begin{figure}[t]
    \centering
    \includegraphics[width=0.7\linewidth]{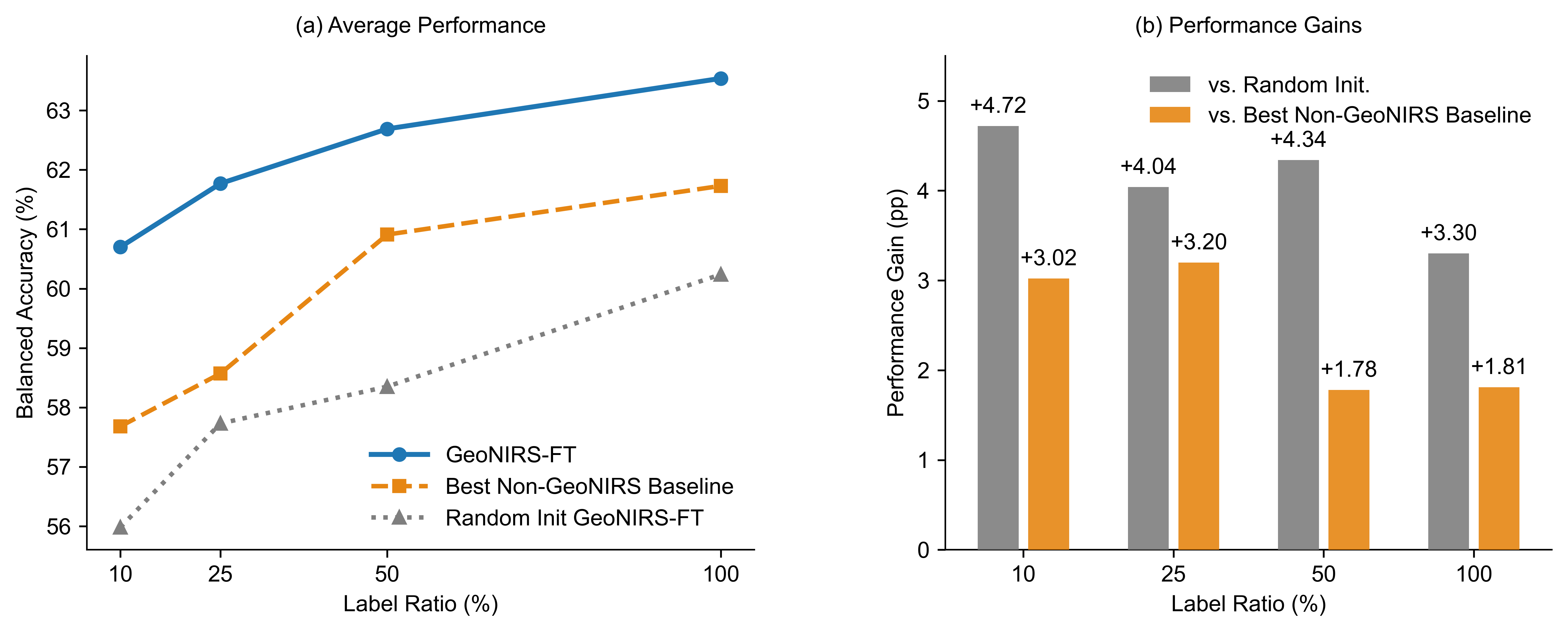}
    \caption{Label-efficient downstream adaptation averaged equally across eight tasks.
    (a) Balanced accuracy of pretrained GeoNIRS-FT, its randomly initialized counterpart, and the strongest non-GeoNIRS baseline at each training-label ratio. The baseline is selected separately at each ratio according to its mean balanced accuracy across the eight tasks.
    (b) Absolute performance gains of GeoNIRS-FT over random initialization and the selected non-GeoNIRS baseline.}
    \label{fig:label_efficiency}
\end{figure}

We next examine whether pretraining improves downstream adaptation when labeled data are scarce. GeoNIRS-FT is trained using class-balanced, nested subsets containing 10\%, 25\%, 50\%, or 100\% of the training data, while the validation and test sets remain unchanged. Figure~\ref{fig:label_efficiency} compares pretrained GeoNIRS-FT with its randomly initialized counterpart and the strongest non-GeoNIRS baseline at each label ratio, selected by mean balanced accuracy across the eight tasks.

GeoNIRS-FT achieves the highest task-averaged accuracy at all four label ratios. Its gain over random initialization is 4.72 percentage points with 10\% labels and 3.30 points with the complete training set. Its gains over the strongest non-GeoNIRS baseline are 3.02, 3.20, 1.78, and 1.81 points at 10\%, 25\%, 50\%, and 100\% labels, respectively. These results indicate that pretraining remains beneficial when downstream supervision is limited.

\subsection{Channel Robustness}
\label{sec:Channel_Robustness}
We evaluate robustness to missing channels by removing 10\%, 25\%, or 50\% of the inputs only during inference. Under fixed channel loss, the same predefined subset of channels is removed in every run; under random channel loss, a new subset is sampled independently for each seed. All models are trained using complete channel inputs.

\begin{figure}[t]
    \centering
    \includegraphics[width=0.7\linewidth]{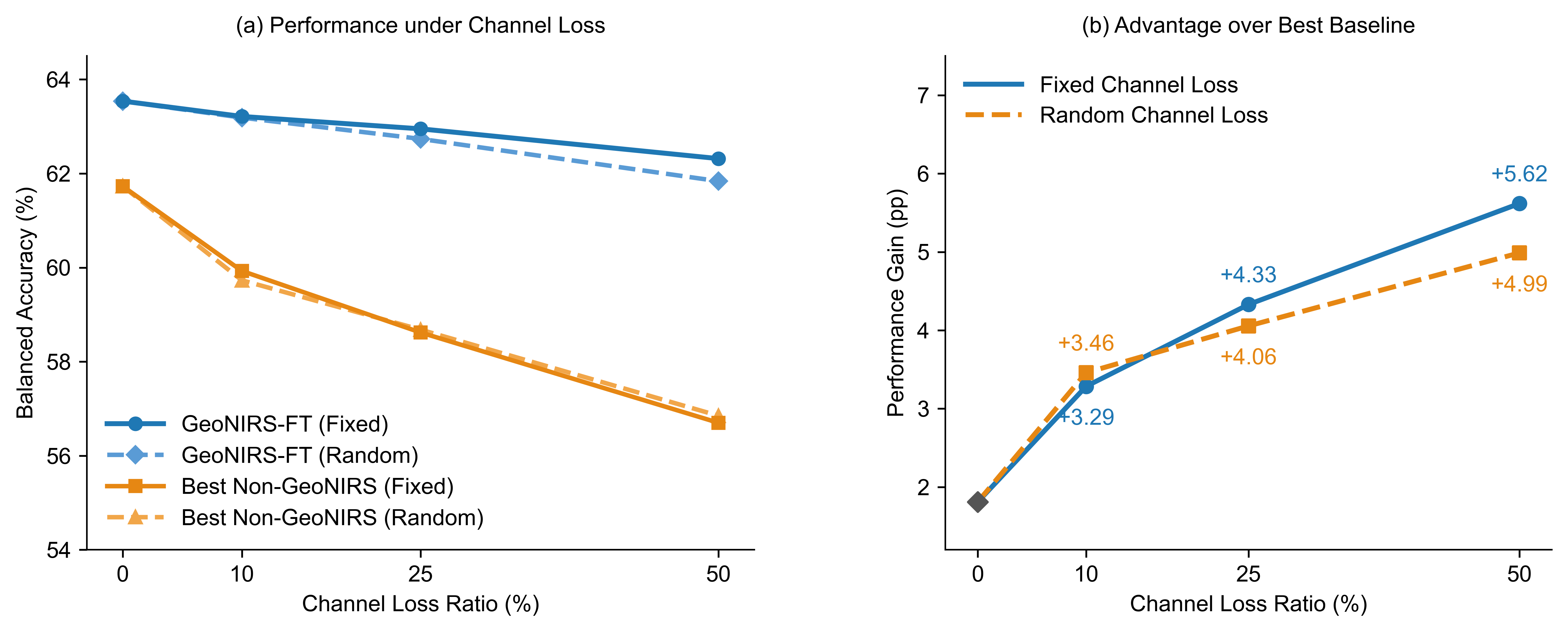}
    \caption{
    Channel-loss robustness averaged equally across eight downstream tasks.
    \textbf{(a)} Balanced accuracy of GeoNIRS-FT and the strongest non-GeoNIRS baseline under fixed and random channel loss. The baseline is selected separately for each condition by its mean balanced accuracy across the eight tasks.
    \textbf{(b)} Absolute performance gains of GeoNIRS-FT over the selected baseline.
    Channel loss is applied only during inference; all models are trained with complete channel inputs.
    }
    \label{fig:channel_robustness}
\end{figure}

As shown in Figure~\ref{fig:channel_robustness}, GeoNIRS-FT remains stable under both settings. At 50\% channel loss, its balanced accuracy decreases by 1.22 points under fixed loss and 1.70 points under random loss. Its advantage over the strongest non-GeoNIRS baseline in each condition increases from 1.81 points with complete channels to 5.62 and 4.99 points under fixed and random 50\% channel loss, respectively. These results are consistent with the geometry-aware design of GeoNIRS, which may help the model use spatially related measurements when some channels are unavailable. Similar trends across full fine-tuning, LoRA, and linear probing are reported in Supplementary Table~\ref{tab:channel_drop_aggregate}.

\subsection{Ablation Studies}
\label{sec:Ablation_Studies}


\paragraph{Geometry perturbation after pretraining.}

We perturb only the channel coordinates during downstream inference while
keeping the pretrained weights and fNIRS signals unchanged. Coordinates are
either replaced with zeros or randomly shuffled across channels. These
perturbations reduce average balanced accuracy by 2.34 and 3.24 percentage
points, respectively, across the three adaptation protocols (Table~\ref{tab:geometry_perturbation}).
The larger reduction under shuffling indicates that incorrect channel--location
correspondence is more harmful than removing geometry entirely.

\begin{table}[t]
\centering
\caption{
Effect of perturbing channel geometry during downstream inference, averaged equally across eight tasks. Results are reported as balanced accuracy (\%), together with the absolute gains in balanced accuracy over competing baselines under geometry perturbation. Standard deviations are obtained by pooling task-level within-task variances.
}
\label{tab:geometry_perturbation}

\small
\setlength{\tabcolsep}{3pt}
\renewcommand{\arraystretch}{1.08}

\begin{tabular*}{\columnwidth}{
@{\extracolsep{\fill}}
lccccc
@{}
}
\toprule
\textbf{Adaptation}
& \textbf{Normal}
& \textbf{Zeros}
& \textbf{Shuffle}
& \makecell{\textbf{Zeros}\\\boldmath$\Delta$ \textbf{(pp)}}
& \makecell{\textbf{Shuffle}\\\boldmath$\Delta$ \textbf{(pp)}} \\
\midrule

Full Fine-Tuning
& $\mathbf{63.54 \pm 6.21}$
& $61.78 \pm 6.72$
& $61.20 \pm 6.05$
& $-1.76$
& $-2.34$ \\

LoRA
& $\mathbf{63.42 \pm 5.30}$
& $60.30 \pm 4.91$
& $59.57 \pm 4.81$
& $-3.12$
& $-3.85$ \\

Linear Probing
& $\mathbf{63.06 \pm 4.96}$
& $60.93 \pm 4.32$
& $59.54 \pm 4.31$
& $-2.13$
& $-3.52$ \\

\midrule

\textbf{Average}
& $\mathbf{63.34 \pm 5.52}$
& $61.00 \pm 5.41$
& $60.10 \pm 5.11$
& $\mathbf{-2.34}$
& $\mathbf{-3.24}$ \\

\bottomrule
\end{tabular*}
\end{table}

\paragraph{Pretraining components.}


We retrain two variants using the same pretraining data and downstream
protocols. The w/o Geometry variant removes channel-coordinate encoding,
whereas Self-Reconstruction masks the same temporal segments in HbO and HbR
and reconstructs each chromophore from its own visible context. As shown in
Table~\ref{tab:pretraining_component_ablation}, the two variants reduce average
balanced accuracy by 2.98 and 1.58 percentage points, respectively, with
consistent degradation across full fine-tuning, LoRA, and linear probing.

\begin{table*}[t]
\centering
\caption{
Ablation study of the proposed pretraining components, averaged equally
across eight downstream tasks. Results are reported as balanced accuracy
(\%). $\Delta$ denotes the change relative to GeoNIRS in percentage points.
}
\label{tab:pretraining_component_ablation}

\footnotesize
\setlength{\tabcolsep}{1.5pt}
\renewcommand{\arraystretch}{1.10}

\begin{tabular*}{\linewidth}{
@{\extracolsep{\fill}}
l*{4}{cc}
@{}
}
\toprule

&
\multicolumn{2}{c}{\textbf{Full Fine-Tuning}}
&
\multicolumn{2}{c}{\textbf{LoRA}}
&
\multicolumn{2}{c}{\textbf{Linear Probing}}
&
\multicolumn{2}{c}{\textbf{Average}}
\\

\cmidrule(lr){2-3}
\cmidrule(lr){4-5}
\cmidrule(lr){6-7}
\cmidrule(lr){8-9}

\textbf{Pretraining Variant}
& \textbf{BA}
& \boldmath$\Delta$
& \textbf{BA}
& \boldmath$\Delta$
& \textbf{BA}
& \boldmath$\Delta$
& \textbf{BA}
& \boldmath$\Delta$
\\

\midrule

GeoNIRS
& \textbf{63.54} & --
& \textbf{63.42} & --
& \textbf{63.06} & --
& \textbf{63.34} & --
\\

w/o Geometry
& 60.93 & $-2.61$
& 60.19 & $-3.23$
& 59.97 & $-3.09$
& 60.36 & $-2.98$
\\

Self-Reconstruction
& 62.29 & $-1.25$
& 61.46 & $-1.96$
& 61.53 & $-1.53$
& 61.76 & $-1.58$
\\

\bottomrule
\end{tabular*}

\vspace{2pt}

\begin{minipage}{0.98\textwidth}
\footnotesize
\end{minipage}

\end{table*}

\section{Conclusion}

We introduced GeoNIRS, a geometry-aware framework for joint pretraining across
fNIRS datasets with heterogeneous channel counts and probe layouts. GeoNIRS
uses source--detector geometry to establish cross-layout spatial correspondence,
compresses variable channel sets into fixed latent representations, and learns
from paired HbO/HbR signals through bidirectional cross-chromophore
reconstruction. Across eight held-out downstream datasets, GeoNIRS achieves
the strongest task-averaged performance among the evaluated methods and
maintains its advantages under linear probing, limited-label adaptation, and
missing-channel conditions. The perturbation and ablation results further
confirm the contributions of channel geometry and cross-chromophore
reconstruction. These findings provide a practical and reproducible foundation
for representation learning across heterogeneous fNIRS datasets.

\newpage

\subsection*{AI use statement}

In this work, we used generative AI tools to assist with the implementation of parts of the proposed method and with translation.
We have not used generative AI tools for designing the proposed methodology or experiments, formulating hypotheses, interpreting experimental results, data cleaning or reformatting, and task related to synthetic data generation, mathematical proofs or conclusions, theoretical or conceptual framework development, and qualitative or thematic analysis are not applicable to this work.
Additionally, we used generative AI tools for general software development assistance, searching for publicly available datasets, and language polishing to improve readability. We have reviewed all AI-assisted work. 
All AI-assisted code and text were manually reviewed and verified by the authors. AI-assisted code used in the experiments was tested for correctness, and datasets suggested through AI-assisted search were verified against their original sources.

\subsection*{Ethics statement}

This work uses publicly available datasets released by previous studies. No new human subjects were recruited, and no additional experiments or data collection were conducted in this work. We only used publicly accessible and anonymized datasets in accordance with their respective data usage policies. This study does not involve the collection or processing of personally identifiable information.

\subsection*{Reproducibility statement}

\section*{Reproducibility Statement}

We provide detailed descriptions of the datasets, preprocessing procedures,
model architecture, training settings, and evaluation protocols in the main
paper and appendix. All datasets used in this work are publicly available.
The complete source code, experiment configurations, and pretrained
checkpoints are publicly available at
\url{https://github.com/GeoNIRS/GeoNIRS}.



\newpage
\bibliography{iclr2027_conference}
\bibliographystyle{iclr2027_conference}

\newpage
\appendix
\section{Tasks and datasets}
\label{app:datasets}



We initially screened nearly 100 publicly available fNIRS datasets and retained 30 that provided the signals and metadata required for standardized preprocessing and geometry-aware modeling. Most excluded datasets lacked source--detector coordinates, raw intensity measurements, or other essential acquisition metadata. Although these datasets may remain suitable for dataset-specific analyses, the missing information prevents their reliable integration into a unified geometry-aware pretraining framework. Among the 30 retained datasets, 22 are used for pretraining and eight are held out exclusively for downstream evaluation. GeoNIRS-base uses the eight pretraining datasets in Type-I(a), whereas GeoNIRS-large additionally uses the 14 datasets in Type-I(b). Type-II lists the eight downstream datasets excluded from both pretraining corpora. The eight-dataset base pretraining corpus contains 42,951 windows.

\vspace{4pt}
\begin{center}
\textit{\textbf{Supplementary Table 1.} Dataset inventory for pretraining and downstream evaluation.}
\end{center}
\vspace{-5pt}
\begingroup
\footnotesize
\setlength{\tabcolsep}{2.4pt}
\renewcommand{\arraystretch}{1.08}
\setlength{\LTpre}{1pt}
\setlength{\LTpost}{0pt}
\providecolor{headergray}{gray}{0.92}
\providecolor{tablecyan}{RGB}{220,238,243}
\arrayrulecolor{black!65}
\begin{xltabular}{\textwidth}{
  >{\RaggedRight\arraybackslash}p{0.160\textwidth}
  >{\RaggedRight\arraybackslash}p{0.090\textwidth}
  >{\Centering\arraybackslash}p{0.080\textwidth}
  >{\Centering\arraybackslash}p{0.095\textwidth}
  >{\Centering\arraybackslash}p{0.090\textwidth}
  >{\RaggedRight\arraybackslash}X }
\toprule
\rowcolor{headergray}
\textbf{Task type} & \textbf{Datasets} & \textbf{\#Subjects} & \textbf{\#Channels} & \textbf{Sample length (s)} & \textbf{Task description} \\
\midrule
\endfirsthead
\toprule
\rowcolor{headergray}
\textbf{Task type} & \textbf{Datasets} & \textbf{\#Subjects} & \textbf{\#Channels} & \textbf{Sample length (s)} & \textbf{Task description} \\
\midrule
\endhead
\midrule
\multicolumn{6}{r}{\textit{Continued on next page}}\\
\endfoot
\bottomrule
\endlastfoot

\rowcolor{tablecyan}
\multicolumn{6}{l}{\textbf{Type-I(a): Base pretraining corpus (8 datasets; also used by Large)}}\\*
\hline
Cognitive task & \cite{anderson2025stroop} & 17 & 214 & 30 & Word-color Stroop with congruent and incongruent conditions; high-density and sparse-array fNIRS acquisition. \\
\hline
Social interaction & \cite{moffat2026intergen} & 122 & 26 & 30 & Dyadic hyperscanning during collaborative drawing with longitudinal follow-up. \\
\hline
Naturalistic stimulation and functional localization & \cite{fogarty2026hddot} & 5 & 3375--3515 & 30 & Audiovisual movie viewing with visual, language, and motor functional localizers. \\
\cline{2-6}
& \cite{fogarty2025vhddot} & 8 & 9543--10759 & 30 & Movie viewing and functional localizers with very-high-density DOT and fMRI comparison. \\
\hline
Resting state & \cite{yucel2025inclusion} & 115 & 26 & 30 & Resting-state recordings with an additional ball task; hair/skin effects on fNIRS signal quality. \\
\cline{2-6}
& \cite{butters2026cambridgerest} & 64 & 1728 & 30 & Resting-state fNIRS recordings. \\
\hline
Sensory stimulation and attention & \cite{bale2026cambridge} & 63 & 432 & 30 & Visual stimulation under two experimental conditions (A/B). \\
\cline{2-6}
& \cite{dubois2024reliability} & 49 & 592--2744 & 30 & Auditory story/noise/silence stimulation and time-domain fNIRS test--retest reliability. \\
\hline

\rowcolor{tablecyan}
\multicolumn{6}{l}{\textbf{Type-I(b): Additional pretraining datasets for Large (14 datasets)}}\\*
\hline
Cognitive task & \cite{ajra2026cmx7mm} & 30 & 16 & 30 & Multimodal recording of low, medium, and high mental workload. \\
\cline{2-6}
& \cite{torrescuevas2025workloaddataset} & 29 & 24 & 30 & Resident-physician shift load at 0 h, 12 h, and 24 h. \\
\hline
Social interaction & \cite{chen2025hyperscanning} & 85\textsuperscript{a} & 20 & 30 & Inter-brain synchrony during social touch in 47 female dyads (24 friend dyads and 23 stranger dyads). \\
\cline{2-6}
& \cite{hu2025appearance} & 31 & 40 & 30 & Natural conversation about physical appearance (NIRx system). \\
\hline
Sensory stimulation and attention & \cite{sanjuan2017tinnitus} & 15 & 44 & 30 & Resting-state functional connectivity in tinnitus (Homer \texttt{.nirs}; approximately 17.6 min per recording). \\
\cline{2-6}
& \cite{duwadi2026cocktail} & 30 & 567 & 30 & Selective auditory attention: attend left vs.\ attend right. \\
\hline
Clinical and disease & \cite{guevara2024parkinson} & 40 & 22 & 30 & Healthy control vs.\ Parkinson's disease during fine motor tasks. \\
\cline{2-6}
& \cite{isaev2024stroke} & 15 & 28 & 30 & Stroke patients controlling a BCI: rest, imagined left hand, and imagined right hand. \\
\cline{2-6}
& \cite{shin2017eegnirs} & 29 & 36 & 30 & Left-hand vs.\ right-hand motor imagery; the source also includes mental arithmetic and baseline recordings. \\
\cline{2-6}
& \cite{sainifarron2026anesthesia} & 8 & 15--20 & 30 & General anesthesia: verbal commands, motor tasks, tetanic stimulation, and loss of consciousness. \\
\hline
Emotion and affect & \cite{goodrick2026yoga} & 37 & 24 & 30 & A 30-day yoga intervention; positive, negative, and neutral affect. \\
\cline{2-6}
& \cite{spape2024nemo} & 31 & 24 & 30 & Emotion perception and affective imagery: positive vs.\ negative valence. \\
\hline
Other & \cite{zhang2025visuomotor} & 49 & 44 & 30 & Visuomotor tasks with simultaneous behaviour, eye tracking, EEG, fNIRS, and EKG; six resting and six task blocks. \\
\cline{2-6}
& \cite{duncan2018cosmetic} & 30 & 52 & 30 & fNIRS during real use of a cosmetic product (ETG-4000); apply, evaluate, type, and baseline epochs. \\
\hline

\rowcolor{tablecyan}
\multicolumn{6}{l}{\textbf{Type-II: Downstream evaluation datasets (8 datasets)}}\\*
\hline
Clinical and disease & \cite{shi2025hefmi} & 37 & 90 & 10 & Healthy control vs.\ intracerebral hemorrhage (ICH), 2 classes. \\
\cline{2-6}
& \cite{balint2025speech} & 72 & 31--39 & 12 & Cochlear-implant users vs.\ normal-hearing participants, 2 classes. \\
\cline{2-6}
& \cite{satish2026tinnitus} & 63 & 83 & 12 & Subjective tinnitus vs.\ healthy control, 2 classes. \\
\hline
Pain and sensory processing & \cite{peng2018morphine} & 14 & 33 & 12 & Electrical-stimulation pain intensity: VAS 3 (low) vs.\ VAS 7 (high), 2 classes. \\
\hline
Emotion and affect & \cite{raible2026mulpa} & 54 & 134 & 12 & Facial emotion perception in MULPA: happy vs.\ sad, 2 classes. \\
\hline
Motor and action observation & \cite{medvedeva2026poststroke} & 16 & 70 & 10 & Movement vs.\ no movement for intact/paretic-hand recordings, 2 classes. \\
\cline{2-6}
& \cite{li2020motorcomplexity} & 21 & 60 & 12 & Action observation vs.\ action execution, 2 classes. \\
\hline
Semantic decoding & \cite{rybar2025semantic} & 12 & 11--14 & 12 & Imagined animal vs.\ imagined tool, 2 classes. \\
\end{xltabular}
\vspace{2pt}
\footnotesize
\noindent\textit{Note:} Subject counts refer to participants included in the pretraining or downstream datasets, not necessarily the full recruited cohorts. Channel counts describe native recordings before quality control and channel selection; ranges indicate variation across recordings.

\smallskip
\noindent\textsuperscript{a}Chen et al. recruited 94 participants in 47 dyads. The recordings included here cover 85 participants: 38 complete dyads and 9 dyads with recordings available for only one partner.
\endgroup

\clearpage

\section{Implementation and Evaluation Details}
\label{app:implementation_details}

\subsection{Signal Preprocessing}
\label{sec:signal_preprocessing}

All datasets are first parsed using format-specific adapters for SNIRF, NIRx,
CSV, BIDS, and other supported formats, and are converted into a unified
recording representation. After format harmonization, both pretraining and
downstream datasets follow the same signal-processing pipeline and
hyperparameters, without dataset-specific preprocessing branches.

\subsubsection{Signal-Level Preprocessing}

\paragraph{Optical density conversion.} For each measurement channel $c$, raw light intensity is converted to optical density (OD) as \begin{equation} \mathrm{OD}_{c}(t) = -\log\left( \frac{I_{c}(t)} {\operatorname{median}_{t'} \left(I_{c}(t')\right)} \right), \end{equation} where the reference baseline is defined as the median of the positive, finite intensity values of that channel within the recording.

\paragraph{Quality control.}
For each measurement channel, robust outlier detection is independently applied
to its optical-density (OD) time series using the median absolute deviation
(MAD). The channel-wise OD median is used as the center, and
$1.4826 \times \mathrm{MAD}$ is used as the robust scale estimate. Time points
with $|z_{\mathrm{robust}}| > 8$ are marked as outliers, and non-finite values
are likewise marked as invalid. Robust $z$-scores are computed only when the
channel MAD is finite and greater than zero. Channels with zero or non-finite
MAD are not subjected to point-wise outlier detection and are subsequently
marked as invalid by the channel-level quality-control criteria.

Contiguous outlier segments are then handled according to their duration.
Segments lasting no more than $2\,\mathrm{s}$ and not touching either boundary
of the recording are linearly interpolated using valid observations from the
same channel, and the interpolated time points are subsequently treated as
valid. Segments lasting longer than $2\,\mathrm{s}$, or extending to either
boundary of the recording, are not interpolated. Their corresponding time
intervals are marked as invalid for the entire recording, and the resulting
temporal-validity mask is shared across all channels.

In addition, a channel is marked as invalid for the entire recording if its MAD
is zero or non-finite, if its post-quality-control standard deviation is no
greater than $10^{-12}$, or if its standard deviation is greater than 10 times
or less than $1/10$ of the median standard deviation across channels in that
recording.

\paragraph{Modified Beer--Lambert law.}
Valid dual-wavelength measurements are converted to oxygenated hemoglobin
(HbO) and deoxygenated hemoglobin (HbR) concentrations using the modified
Beer--Lambert law (MBLL), with the corresponding source--detector distance and
a fixed partial pathlength factor of
\begin{equation}
    \mathrm{PPF}=6.0.
\end{equation}
The resulting concentrations are expressed in $\mu\mathrm{mol/L}$. A
consistent physical channel ordering and the corresponding channel-validity
mask are established during this step.

\paragraph{Resampling}

HbO and HbR signals are separately resampled to 5~Hz only over valid time segments. Validity is determined before computing the two chromophore signals: temporal validity is defined by the quality-control procedure, while channel validity is determined by the channel-level criteria. A signal position is considered valid only when both the corresponding time point and channel are valid.

\paragraph{Normalization.}
HbO and HbR are standardized separately for each recording using
\begin{equation}
    x'
    =
    \frac{x-\mu}{\sigma}.
\end{equation}
For each signal stream, a single recording-level mean $\mu$ and standard
deviation $\sigma$ are estimated from all valid channels and valid time points.
Unlike channel-wise normalization, this preserves relative amplitude
differences across channels while reducing recording-level scale variation.

\subsubsection{Sample Construction}

For pretraining, all samples are constructed using fixed 30~s windows, corresponding to 150 time points at a sampling rate of 5~Hz. Each window is further divided into 15 consecutive 2~s segments, with 10 time points per segment.

For downstream tasks, we preserve the temporal window length defined by each original task and do not apply additional temporal padding, cropping, or re-segmentation. Specifically, \cite{shi2025hefmi} and \cite{medvedeva2026poststroke} use 10~s windows, while the remaining six downstream tasks use 12~s windows.

During pretraining, a 2~s segment participates in masking and cross-modal reconstruction loss computation only when at least 80\% of its time points are valid. This prevents heavily corrupted temporal segments from affecting the self-supervised learning objective.

For downstream evaluation, the channel set is fixed separately for each dataset. We retain the channels that are jointly available across all quality-controlled recordings included in that dataset, and reconstruct all samples using exactly the same channel set and channel ordering. This design is both necessary and fair: baselines with fixed input dimensionality cannot accommodate recording-specific channel counts, while using a common channel set ensures that all subjects, recordings, and compared models receive identical channel inputs. Consequently, recording-level variation in channel availability cannot serve as a potential subject- or class-specific cue. Invalid samples along the temporal dimension are zero-filled and indicated by the corresponding validity mask.

\subsubsection{Channel Sampling for Ultra-High-Density Pretraining Data}

  To preserve heterogeneous spatial structure while keeping the computational cost
  of ultra-high-density recordings tractable, we cap the number of physical
  channels entering the channel-wise attention module at 512 per pretraining
  window. In our pretraining corpus this cap is reached by five datasets:
  \cite{fogarty2025vhddot} (9{,}543--10{,}759 channels),
  \cite{fogarty2026hddot} (3{,}375--3{,}515),
  \cite{butters2026cambridgerest} (1{,}728),
  \cite{dubois2024reliability} (592--2{,}744), and
  \cite{duwadi2026cocktail} (567 channels).

  For these recordings we do not use a fixed subset of 512 channels throughout
  pretraining. Instead, whenever the number of valid channels in a window exceeds
  the cap, we sample \emph{exactly} 512 channels without replacement from that
  window's valid channel set; windows at or below the cap keep all of their valid
  channels. The sampling seed is a deterministic function of the run seed, the
  recording identifier, the window start, and the training epoch, so the selection
  is reproducible given the configuration while different channel subsets are
  observed across epochs. For validation windows the epoch is replaced by a fixed
  constant, so the same subset is used at every validation pass and validation
  losses remain comparable across epochs.

\subsection{Channel Geometry Representation}
\label{sec:channel_geometry}

Each physical channel $c$ is represented by a six-dimensional geometry vector,
\begin{equation}
    \mathbf{g}_c
    =
    \left[
        \mathrm{mid}_{c,x},
        \mathrm{mid}_{c,y},
        \mathrm{mid}_{c,z},
        \Delta_{c,x},
        \Delta_{c,y},
        \Delta_{c,z}
    \right],
\end{equation}
where
\begin{equation}
    \mathbf{mid}_c
    =
    \frac{\mathbf{s}_c+\mathbf{d}_c}{2},
    \qquad
    \boldsymbol{\Delta}_c
    =
    \mathbf{d}_c-\mathbf{s}_c,
\end{equation}
and $\mathbf{s}_c$ and $\mathbf{d}_c$ denote the coordinates of the source
and detector associated with channel $c$, respectively. The midpoint
$\mathbf{mid}_c$ describes the spatial location of the channel, whereas
$\boldsymbol{\Delta}_c$ captures the source--detector separation and
orientation. The resulting representation therefore jointly encodes channel
position and source--detector geometry.

\paragraph{Coordinate harmonization and normalization.}
Coordinate units vary across datasets. Across the 30 datasets used in this
work (22 for pretraining and 8 for downstream evaluation), the original
source and detector coordinates are represented in millimeters, centimeters,
or meters. We first convert all coordinates to a common physical unit and
then normalize them using a fixed reference scale of 25.46~mm. The resulting
geometry features are dimensionless while preserving the relative spatial
organization and source--detector configuration of the original probe layout.

Geometry coordinates are additionally centered to remove global translation
differences while preserving relative channel positions and source--detector
orientations. During pretraining, centering is performed independently for
each recording using its valid source and detector coordinates. For downstream
evaluation, a fixed dataset-level geometry representation is constructed after
the common physical channel set has been determined and is shared by all
samples within that dataset. Consequently, geometry does not vary across
samples, subjects, or class labels within the same downstream task and
provides only the fixed spatial structure of the corresponding probe layout.

\paragraph{Planar probe layouts.}
Not all datasets provide fully three-dimensional source and detector
coordinates. Among the eight downstream datasets, three use planar probe
layouts: \cite{shi2025hefmi}, \cite{peng2018morphine}, and
\cite{li2020motorcomplexity}. Among the 22 pretraining datasets, six use
planar layouts: \cite{zhang2025visuomotor},
\cite{sanjuan2017tinnitus}, \cite{spape2024nemo},
\cite{guevara2024parkinson}, \cite{goodrick2026yoga}, and
\cite{torrescuevas2025workloaddataset}.

Together, these six planar pretraining datasets account for 161.76~h of data,
corresponding to 19.6\% of the full 827.04-h pretraining corpus. Depending on
the original dataset, planar geometry is either provided directly as
two-dimensional coordinates or stored in a three-dimensional representation
with all source and detector $z$ coordinates equal to zero, as in
\cite{shi2025hefmi}.

For planar layouts, $\mathrm{mid}_{c,z}$ and $\Delta_{c,z}$ are therefore
identically zero, while the remaining four dimensions continue to encode the
in-plane channel position, source--detector separation, and orientation. We
retain the same six-dimensional representation without introducing synthetic
depth coordinates, allowing planar and three-dimensional probe layouts to be
processed by the same geometry encoder.

\subsection{Model Configurations}
\label{sec:model_configurations}

\begin{table}[t]
\centering
\caption{Backbone configurations of GeoNIRS-base and GeoNIRS-large.}
\label{tab:model_configurations}
\small
\setlength{\tabcolsep}{5pt}
\renewcommand{\arraystretch}{1.12}
\begin{tabular}{lcc}
\toprule
\textbf{Component} &
\textbf{GeoNIRS-base} &
\textbf{GeoNIRS-large} \\
\midrule
Backbone parameters
    & 6,511,214
    & 30,146,414 \\

Embedding width
    & 192
    & 384 \\

Temporal Transformer layers / heads
    & 4 / 8
    & 6 / 12 \\

Projection dimensions
    & $[192,128,96,64]$
    & $[384,192,96,128]$ \\

Output feature dimension $F$
    & 64
    & 128 \\

Geometry encoder
    & $6\!\rightarrow\!192$ / 8 heads / 1 layer
    & $6\!\rightarrow\!384$ / 12 heads / 1 layer \\

Base representation shape
    & $[B,2,16,P,64]$
    & $[B,2,16,P,128]$ \\
\bottomrule
\end{tabular}
\end{table}

\subsection{Pretraining Details}
\label{sec:pretraining_details}

Both backbone scales use the same optimization framework, with
scale-specific differences only in the training budget, global batch size,
peak learning rate, and numerical precision. The shared and scale-specific
pretraining configurations are summarized in
Table~\ref{tab:pretraining_details}.

\begin{table}[t]
\centering
\caption{Pretraining configurations for GeoNIRS-6.5M and GeoNIRS-30M.
Shared settings are reported once across both model scales.}
\label{tab:pretraining_details}
\small
\setlength{\tabcolsep}{5pt}
\renewcommand{\arraystretch}{1.12}
\begin{tabular}{lcc}
\toprule
\textbf{Hyperparameter} &
\textbf{GeoNIRS-6.5M} &
\textbf{GeoNIRS-30M} \\
\midrule

\multicolumn{3}{l}{\textit{Shared optimization settings}} \\
Optimizer
    & \multicolumn{2}{c}{AdamW} \\
Weight decay
    & \multicolumn{2}{c}{0.05} \\
Adam betas
    & \multicolumn{2}{c}{$(0.9,\,0.95)$} \\
Adam epsilon
    & \multicolumn{2}{c}{$1\times10^{-6}$} \\
Gradient clipping
    & \multicolumn{2}{c}{1.0} \\
Learning-rate schedule
    & \multicolumn{2}{c}{10\% warmup + cosine decay} \\
Minimum learning-rate ratio
    & \multicolumn{2}{c}{0.1} \\
Distributed training
    & \multicolumn{2}{c}{4-GPU DDP} \\
Random seed
    & \multicolumn{2}{c}{810} \\

\midrule
\multicolumn{3}{l}{\textit{Scale-specific settings}} \\

Peak learning rate
    & $1\times10^{-4}$
    & $3\times10^{-4}$ \\

Training epochs
    & 150 
    & 40 \\

Global batch size
    & 48 (12/GPU $\times$ 4)
    & 96 (24/GPU $\times$ 4) \\

\bottomrule
\end{tabular}
\end{table}

\subsection{Downstream Adaptation}
\label{sec:downstream_adaptation}

\subsubsection{Unified Training Protocol}

All neural methods use identical downstream samples, fold assignments, channel
ordering, labels, and the optimization protocol summarized in
Table~\ref{tab:downstream_protocol}.

\begin{table}[t]
\centering
\caption{Unified downstream training protocol.}
\label{tab:downstream_protocol}
\small
\setlength{\tabcolsep}{6pt}
\renewcommand{\arraystretch}{1.10}
\begin{tabular}{lc}
\toprule
\textbf{Hyperparameter} & \textbf{Setting} \\
\midrule
Epochs & 30 \\
Batch size & 16 \\
Optimizer & AdamW \\
Weight decay & 0.05 \\
Gradient clipping & 1.0 \\
Warmup & 3 epochs \\
Schedule & Cosine decay to $1\times10^{-6}$ \\
Loss & Class-weighted cross-entropy \\
Early-stopping patience & 15 \\
Model selection & Validation balanced accuracy \\
Test evaluation & Once after model selection \\
\bottomrule
\end{tabular}
\end{table}

\subsubsection{GeoNIRS Adaptation}

The backbone representation
$\mathbf{Z}\in\mathbb{R}^{B\times2\times16\times P\times F}$
is flattened, standardized using training-set statistics, followed by dropout
($p=0.1$) and a single linear classification layer. No hidden layer or
LayerNorm is used. Table~\ref{tab:geonirs_adaptation} summarizes the trainable
components and learning rates used for the three adaptation protocols.

\begin{table}[t]
\centering
\caption{GeoNIRS downstream adaptation modes.}
\label{tab:geonirs_adaptation}
\small
\setlength{\tabcolsep}{5pt}
\renewcommand{\arraystretch}{1.10}
\begin{tabular}{lll}
\toprule
\textbf{Mode} &
\textbf{Trainable part} &
\textbf{Learning rate} \\
\midrule
\texttt{GeoNIRS\_lp}
& Classification head 
& $1\times10^{-4}$ \\

\texttt{GeoNIRS\_lora}
& Head + LoRA 
& $1\times10^{-4}$ \\

\texttt{GeoNIRS\_ft}
& Head + backbone
& Head: $10^{-4}$; backbone: $10^{-5}$ \\
\bottomrule
\end{tabular}
\end{table}

For LoRA adaptation, we use rank $r=4$ and LoRA scaling parameter $\alpha=8$,
corresponding to a scaling factor of $\alpha/r=2$, with no adapter dropout.
LoRA adapters are applied to the linear projections in the downstream encoder
path, including the HbO/HbR temporal encoders, geometry encoder,
geometry-aware channel compressor, and chromophore feature projections.
All original backbone parameters remain frozen, and only the LoRA parameters
and task-specific classification head are optimized. Both use a learning rate
of $1\times10^{-4}$. The LoRA down-projection is initialized with
Kaiming-uniform initialization and the up-projection with zeros.

The pretrained reconstruction decoders are removed and are not used during
downstream adaptation.

\subsubsection{Baselines and Controls}

Baseline models receive the same frozen downstream inputs as GeoNIRS.
Input adapters are restricted to tensor reshaping or permutation and do not
perform model-specific normalization, interpolation, channel selection, or
sample filtering. Neural baselines follow the unified protocol in
Table~\ref{tab:downstream_protocol}; model-specific adaptation follows the
corresponding pretrained or randomly initialized setting.

Parameter counts are summarized in
Table~\ref{tab:baseline_parameters}.

\begin{table}[H]
    \centering
    \caption{Parameter counts of the downstream models. All counts are
    reported at the \cite{medvedeva2026poststroke} geometry (52 fNIRS
    channels,
    50 time points), and include the task-level classifier head.}
    \label{tab:baseline_parameters}
    \small
    \setlength{\tabcolsep}{10pt}
    \renewcommand{\arraystretch}{1.10}
    \begin{tabular}{lc}
    \toprule
    \textbf{Model} & \textbf{Parameters} \\
    \midrule
    fNIRS-T & 1.70M \\
    fNIRS4D & 164.4K \\
    fNIRSNet & 0.72K \\
    1D-CNN & 88.4K \\
    EEGNet & 1.52K \\
    Jung-MAE  & 2.03M \\
    MOMENT-base & 109.7M \\
    \midrule
    GeoNIRS-base & 6.51M \\
    GeoNIRS-large & 30.15M \\
    \midrule
    Stat-LDA & Non-neural \\
    RBF-SVM & Non-neural \\
    \bottomrule
    \end{tabular}
    
  \end{table}

\begin{table}[H]
\centering
\caption{Protocol-level control experiments.}
\label{tab:protocol_controls}
\small
\setlength{\tabcolsep}{5pt}
\renewcommand{\arraystretch}{1.10}
\begin{tabular}{p{0.22\textwidth}p{0.70\textwidth}}
\toprule
\textbf{Control} & \textbf{Modification} \\
\midrule
Random initialization
& Replace the pretrained checkpoint with random initialization. \\

Label efficiency
& Use nested class-balanced 10\%, 25\%, and 50\% training subsets. \\

Channel dropout
& Zero and invalidate 10\%, 25\%, or 50\% of channels at test time. \\

Geometry ablation
& Zero or permute channel geometry while keeping signals unchanged. \\
\bottomrule
\end{tabular}
\end{table}

\subsection{Evaluation and Aggregation}
\label{sec:evaluation}

\subsubsection{Evaluation Protocol}

We use five-fold cross-validation with two random seeds (810 and 811) per
fold. An evaluation cell is defined by a (dataset, fold, seed) triplet.
Within each cell, all methods receive identical sample rows, data splits,
labels, and channel ordering. Sample identity, split disjointness, and input
consistency are verified before training.

The benchmark consists of the eight downstream tasks listed in
Table~\ref{tab:benchmark_tasks}. Channel counts correspond to the frozen
common-channel tensors actually used for evaluation, rather than the native
channel counts of the original recordings.

\begin{table}[H]
\centering
\caption{Downstream benchmark tasks and frozen input channel counts.}
\label{tab:benchmark_tasks}
\small
\setlength{\tabcolsep}{6pt}
\renewcommand{\arraystretch}{1.10}
\begin{tabular}{llc}
\toprule
\textbf{Dataset} & \textbf{Task} & \textbf{\#Channels} \\
\midrule
DS7   & Intracerebral hemorrhage & 18 \\
DS47  & Cochlear-implant status  & 10 \\
DS36  & Tinnitus                 & 28 \\
DS19  & Pain intensity           & 23 \\
DS70E & Facial emotion           & 66 \\
DS2   & Semantic category        & 6  \\
DS8   & Motor recovery           & 52 \\
DS42  & Action mode              & 41 \\
\bottomrule
\end{tabular}
\end{table}

Balanced accuracy is the primary metric, with macro-F1 and AUROC reported as
additional metrics. Neural checkpoints are selected by validation balanced
accuracy, and the test split is evaluated once using the selected checkpoint.

\subsubsection{Aggregation and Statistics}

For each method and task, scores are averaged over the ten evaluation cells
(5 folds $\times$ 2 seeds). The reported \textit{Average} is the unweighted
mean across the eight downstream tasks, so each task contributes equally
regardless of its sample size or channel count.

Ablation effects are computed within matched (dataset, fold, seed) cells before
aggregation, ensuring paired comparisons under identical data partitions and
random seeds.

Unless otherwise stated, ``$\pm$'' denotes the standard deviation across the
ten cells for each method--task pair, combining fold and seed variability.
For model-scaling summaries, ``$\pm$'' denotes variability across downstream
tasks after matched comparisons between model scales.

\section{Comparison with Existing fNIRS Pretraining Frameworks}
\label{app:framework_comparison}

Table~\ref{tab:pretraining_framework_comparison} compares existing fNIRS
pretraining frameworks. Heterogeneous channel configurations refer to
differences in channel number or probe layout across pretraining datasets.
Explicit heterogeneity handling requires a dedicated mechanism for resolving
cross-dataset channel correspondence, rather than merely accepting different
sequence lengths. Public checkpoints refer to downloadable pretrained
model weights.

\begin{table}[H]
\centering

\caption{Comparison of fNIRS pretraining frameworks.}

\label{tab:pretraining_framework_comparison}

\scriptsize
\setlength{\tabcolsep}{0.5pt}
\renewcommand{\arraystretch}{1.08}

\begin{tabular*}{\linewidth}{
@{\extracolsep{\fill}}
lcccccc
@{}
}
\toprule
\textbf{Method}
& \makecell{\textbf{Multi-dataset}\\\textbf{pretraining}}
& \makecell{\textbf{Heterogeneous}\\\textbf{channels}}
& \makecell{\textbf{Explicit heterogeneity}\\\textbf{handling}}
& \makecell{\textbf{Unseen-task}\\\textbf{transfer}}
& \makecell{\textbf{Public}\\\textbf{code}}
& \makecell{\textbf{Public}\\\textbf{checkpoint}} \\
\midrule

\citet{liu2021unsupervised}
& $\times$ & $\times$ & $\times$
& $\times$ & $\times$ & $\times$ \\

\citet{wang2021taming}$^{a}$
& $\times$ & $\times$ & $\times$
& $\times$ & $\checkmark$ & $\times$ \\

NIRS-X~\citep{wang2024empower}$^{b}$
& $\times$ & $\times$ & $\times$
& $\times$ & $\checkmark$ & $\times$ \\

\citet{jung2025fnirs}$^{c}$
& $\checkmark$ & $\checkmark$ & $\times$
& $\checkmark$ & $\times$ & $\times$ \\

EFRM~\citep{jung2025efrm}$^{d}$
& $\times$ & $\times$ & $\times$
& $\checkmark$ & $\checkmark$ & $\times$ \\

\citet{subedi2026interpretable}$^{e}$
& $\checkmark$ & $\times$ & $\triangle$
& $\checkmark$ & $\checkmark$ & $\times$ \\

ASDGPrompt~\citep{cai2026asdgprompt}
& $\times$ & $\times$ & $\times$
& $\times$ & $\times$ & $\times$ \\

DSSER~\citep{wang2026dsser}
& $\times$ & $\times$ & $\times$
& $\times$ & $\times$ & $\times$ \\

\midrule
\textbf{GeoNIRS (Ours)}
& $\checkmark$ & $\checkmark$ & $\checkmark$
& $\checkmark$ & $\checkmark$ & $\checkmark$ \\

\bottomrule
\end{tabular*}
\vspace{2pt}

\begin{minipage}{\linewidth}
\scriptsize
\textit{Notes.}
$^{a}$Datasets are used in separate experiments, not joint pretraining.
$^{b}$The released implementation selects one dataset at a time.
$^{c}$Different channel counts are supported without modeling physical
channel coordinates or cross-layout correspondence.
$^{d}$EFRM uses one fNIRS dataset together with EEG data.
$^{e}$Pretraining datasets share a probe configuration; transfer to a new
configuration requires a supervised adapter. $\triangle$ denotes partial support.
\end{minipage}

\end{table}

\section{Additional Experimental Results}
\label{app:additional_results}

\subsection{Cross-Task Performance under Additional Metrics}

To complement the balanced-accuracy results reported in the main text, Tables~\ref{tab:benchmark_macro_f1} and~\ref{tab:benchmark_auroc} present cross-task performance measured by macro F1 and AUROC, respectively. GeoNIRS achieves the strongest task-averaged performance under both additional metrics, supporting the overall conclusions of the main benchmark.

\begin{table}[H]
\centering
\caption{
Cross-task fNIRS decoding performance measured by macro F1 (\%).
Results are averaged over two random seeds, with subject-level five-fold
cross-validation performed for each seed. The best and second-best results
in each column are highlighted in bold and underline, respectively.
}
\label{tab:benchmark_macro_f1}

\setlength{\tabcolsep}{1.5pt}
\renewcommand{\arraystretch}{1.08}
\footnotesize

\begin{tabular*}{\linewidth}{
@{\extracolsep{\fill}}
l*{9}{c}
@{}
}
\toprule

\textbf{Method}
& \textbf{ICH}
& \makecell{\textbf{CI}\\\textbf{Status}}
& \textbf{Tinnitus}
& \makecell{\textbf{Pain}\\\textbf{Intensity}}
& \textbf{Emotion}
& \textbf{Semantic}
& \makecell{\textbf{Motor}\\\textbf{Recovery}}
& \makecell{\textbf{Action}\\\textbf{Mode}}
& \textbf{Average}
\\

\midrule

RBF-SVM~\citeyearpar{naseer2016classification}
& 73.89
& 63.86
& 52.74
& 51.89
& 55.52
& 48.77
& 57.63
& 59.37
& 57.96
\\

Stat-LDA~\citeyearpar{naseer2016lda}
& 51.25
& 51.26
& 48.33
& 53.89
& \textbf{59.18}
& 50.08
& 62.32
& 65.78
& 55.26
\\

\midrule

1D-CNN~\citeyearpar{trakoolwilaiwan2018cnn}
& 64.87
& 56.69
& 54.60
& 50.05
& 57.94
& 48.60
& 55.50
& 55.53
& 55.47
\\

EEGNet~\citeyearpar{lawhern2018eegnet}
& 55.67
& 46.49
& 50.00
& 49.50
& 51.69
& 48.02
& 59.59
& 64.59
& 53.19
\\

fNIRS-T~\citeyearpar{wang2022transformer}
& 70.85
& 63.66
& 49.52
& 51.05
& 54.00
& 48.57
& \underline{64.22}
& \underline{70.72}
& 59.07
\\

fNIRSNet~\citeyearpar{wang2023fnirsnet}
& 47.33
& 38.10
& 32.79
& 33.33
& 33.93
& 33.34
& 55.02
& 33.33
& 38.40
\\

fNIRS4D~\citeyearpar{wang2026fnirs4d}
& 74.62
& 51.22
& 47.38
& 50.79
& 54.85
& 49.56
& 57.67
& 68.98
& 56.88
\\

\midrule

MOMENT-LP~\citeyearpar{goswami2024moment}
& 73.33
& 59.29
& 52.81
& 52.45
& 53.68
& 45.09
& 58.87
& 69.74
& 58.16
\\

MOMENT-LoRA~\citeyearpar{goswami2024moment}
& 71.70
& 61.37
& 51.75
& 52.42
& 53.71
& 47.10
& 62.37
& 70.60
& 58.88
\\

\midrule

Jung-MAE-LP~\citeyearpar{jung2025fnirs}
& 70.95
& 63.65
& 50.44
& 54.23
& 57.74
& 50.55
& 61.34
& 68.36
& 59.66
\\

Jung-MAE-LoRA~\citeyearpar{jung2025fnirs}
& 76.96
& 63.37
& 53.09
& 53.62
& 56.83
& 49.93
& 63.38
& 70.05
& 60.90
\\

\midrule

GeoNIRS-LP (ours)
& 77.08
& 65.19
& \underline{57.80}
& \textbf{55.98}
& 58.15
& \underline{50.60}
& 64.08
& 70.18
& 62.38
\\

GeoNIRS-LoRA (ours)
& \underline{79.57}
& \underline{68.22}
& \textbf{57.87}
& 54.19
& \underline{59.04}
& \textbf{51.01}
& 63.76
& 70.13
& \underline{62.97}
\\

GeoNIRS-FT (ours)
& \textbf{80.29}
& \textbf{68.80}
& 56.74
& \underline{54.36}
& 58.69
& 50.50
& \textbf{64.27}
& \textbf{70.85}
& \textbf{63.06}
\\

\bottomrule
\end{tabular*}

\end{table}

\begin{table}[H]
\centering
\caption{
Cross-task fNIRS decoding performance measured by AUROC (\%).
Results are averaged over two random seeds, with subject-level five-fold
cross-validation performed for each seed. The best and second-best results
in each column are highlighted in bold and underline, respectively.
}
\label{tab:benchmark_auroc}

\setlength{\tabcolsep}{1.5pt}
\renewcommand{\arraystretch}{1.08}
\footnotesize

\begin{tabular*}{\linewidth}{
@{\extracolsep{\fill}}
l*{9}{c}
@{}
}
\toprule

\textbf{Method}
& \textbf{ICH}
& \makecell{\textbf{CI}\\\textbf{Status}}
& \textbf{Tinnitus}
& \makecell{\textbf{Pain}\\\textbf{Intensity}}
& \textbf{Emotion}
& \textbf{Semantic}
& \makecell{\textbf{Motor}\\\textbf{Recovery}}
& \makecell{\textbf{Action}\\\textbf{Mode}}
& \textbf{Average}
\\

\midrule

RBF-SVM~\citeyearpar{naseer2016classification}
& 82.22
& 74.42
& 56.45
& 55.99
& 59.09
& 50.34
& 61.03
& 65.06
& 63.07
\\

Stat-LDA~\citeyearpar{naseer2016lda}
& 53.77
& 50.45
& 52.24
& 55.82
& 61.83
& 50.83
& 67.02
& 72.38
& 58.04
\\

\midrule

1D-CNN~\citeyearpar{trakoolwilaiwan2018cnn}
& 74.04
& 62.42
& 60.67
& 50.76
& 59.64
& 50.97
& 57.74
& 59.67
& 59.49
\\

EEGNet~\citeyearpar{lawhern2018eegnet}
& 59.31
& 49.80
& 49.92
& 50.55
& 53.06
& 50.16
& 63.43
& 70.92
& 55.89
\\

fNIRS-T~\citeyearpar{wang2022transformer}
& 80.76
& 70.46
& 54.34
& 56.34
& 58.45
& \textbf{52.05}
& 70.33
& 77.10
& 64.98
\\

fNIRSNet~\citeyearpar{wang2023fnirsnet}
& 53.29
& 48.43
& 48.72
& 51.54
& 49.04
& 50.12
& 57.36
& 50.71
& 51.15
\\

fNIRS4D~\citeyearpar{wang2026fnirs4d}
& 85.42
& 51.82
& 51.14
& 52.07
& 56.35
& 50.50
& 61.03
& 75.91
& 60.53
\\

\midrule

MOMENT-LP~\citeyearpar{goswami2024moment}
& 80.56
& 65.26
& 55.12
& \underline{57.02}
& 59.07
& 50.09
& \underline{71.26}
& 77.91
& 64.53
\\

MOMENT-LoRA~\citeyearpar{goswami2024moment}
& 78.19
& 68.01
& 55.42
& 56.37
& 57.36
& \underline{51.53}
& 70.08
& \textbf{80.28}
& 64.66
\\

\midrule

Jung-MAE-LP~\citeyearpar{jung2025fnirs}
& 85.04
& 72.60
& 52.01
& \textbf{58.39}
& \textbf{63.13}
& 50.69
& 67.27
& 76.78
& 65.74
\\

Jung-MAE-LoRA~\citeyearpar{jung2025fnirs}
& \underline{89.32}
& 71.13
& 54.46
& 56.74
& 61.86
& 50.41
& 69.96
& 77.32
& 66.40
\\

\midrule

GeoNIRS-LP (ours)
& 86.96
& 75.73
& \textbf{62.56}
& 56.83
& 62.37
& 51.37
& 69.38
& 78.40
& \underline{67.95}
\\

GeoNIRS-LoRA (ours)
& 88.19
& \underline{76.20}
& \underline{62.26}
& 55.94
& 62.08
& 51.00
& 68.89
& 78.82
& 67.92
\\

GeoNIRS-FT (ours)
& \textbf{89.63}
& \textbf{76.40}
& 60.86
& 55.47
& \underline{62.68}
& 50.79
& \textbf{71.29}
& \underline{79.04}
& \textbf{68.27}
\\

\bottomrule
\end{tabular*}

\end{table}

\clearpage

\begin{figure}[H]
    \centering
    \includegraphics[width=0.8\linewidth]{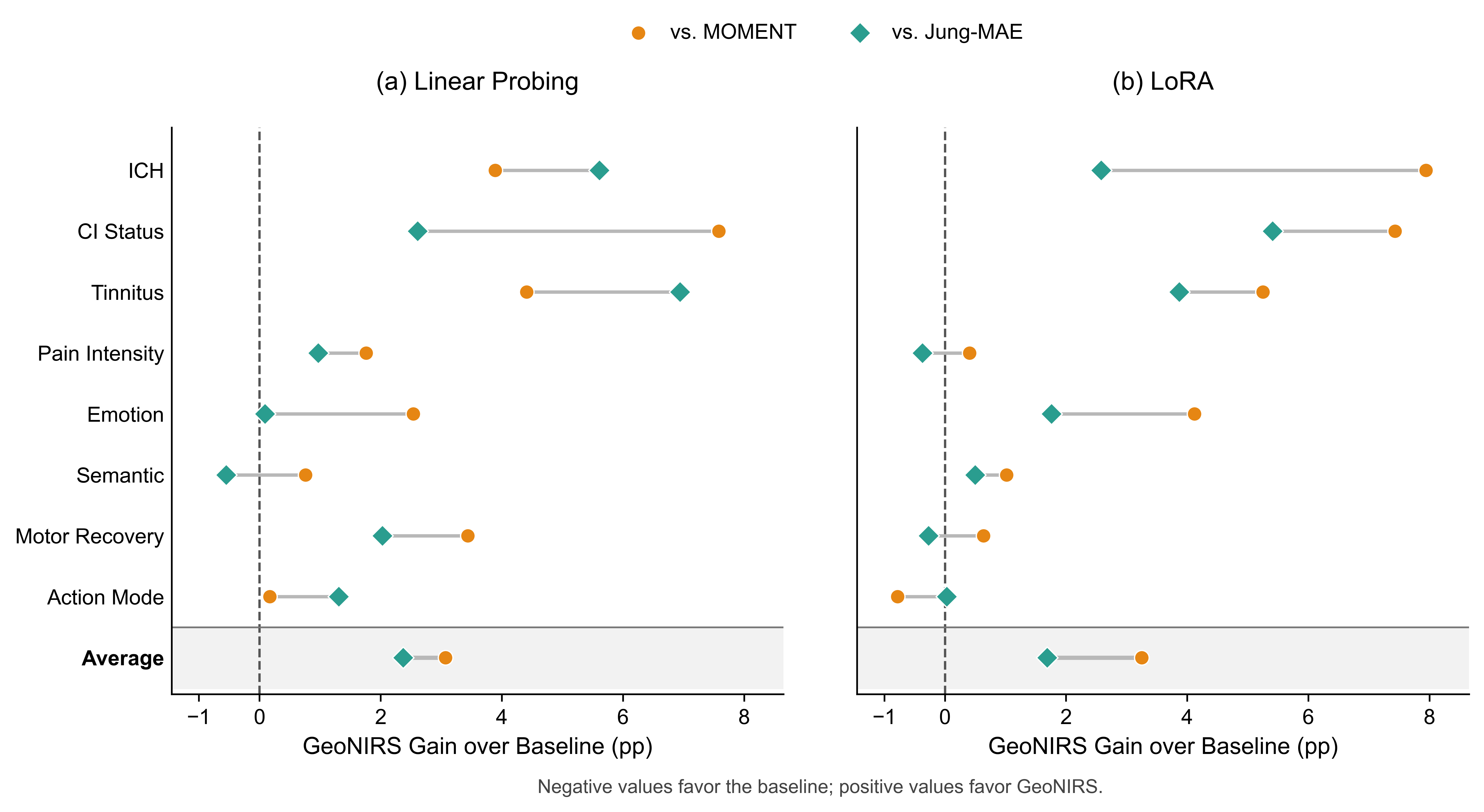}
    \caption{
    Task-wise comparison with pretrained baselines under \textbf{(a)} linear probing and \textbf{(b)} LoRA. Gains are calculated as the balanced accuracy of GeoNIRS minus that of MOMENT or Jung-MAE in percentage points. Orange circles and teal diamonds denote the two baseline comparisons, respectively, while gray lines connect results for the same downstream task. Positive values favor GeoNIRS.
    }
    \label{fig:pretrained_baseline_gains}
\end{figure}

To provide a more direct comparison with existing pretrained models, Figure~\ref{fig:pretrained_baseline_gains} visualizes the task-wise balanced-accuracy differences between GeoNIRS and the matched MOMENT and Jung-MAE baselines under linear probing and LoRA. GeoNIRS achieves positive average margins over both baselines under both adaptation protocols, with most task-level comparisons also favoring GeoNIRS. In particular, the average gains are 3.07 and 2.38 percentage points over MOMENT and Jung-MAE under linear probing, and 3.25 and 1.69 points under LoRA, respectively.

\subsection{Label-Efficient Adaptation}

Table~\ref{tab:label_efficiency} compares pretrained GeoNIRS-FT with
its randomly initialized counterpart and the strongest non-GeoNIRS model
at each label ratio, selected according to its mean balanced accuracy
across the eight tasks. The 10\%, 25\%, and 50\% training subsets are
class-balanced and nested, with the validation and test sets unchanged.

\begin{table}[H]
\centering
\caption{
Label-efficient downstream adaptation measured by balanced accuracy (\%),
averaged equally across eight tasks. Results at 100\% labels are taken from
the main benchmark.
}
\label{tab:label_efficiency}

\setlength{\tabcolsep}{6pt}
\renewcommand{\arraystretch}{1.08}
\footnotesize

\begin{tabular*}{\linewidth}{
@{\extracolsep{\fill}}
lcccc
@{}
}
\toprule

\textbf{Method}
& \textbf{10\%}
& \textbf{25\%}
& \textbf{50\%}
& \textbf{100\%}
\\

\midrule

GeoNIRS-FT
& \textbf{60.70}
& \textbf{61.77}
& \textbf{62.69}
& \textbf{63.54}
\\

GeoNIRS-LoRA
& \underline{60.50}
& \underline{61.44}
& \underline{62.06}
& \underline{63.42}
\\

GeoNIRS-LP
& 60.19
& 60.91
& 61.87
& 63.06
\\

\midrule

GeoNIRS-FT (Random Init)
& 55.98
& 57.73
& 58.35
& 60.25
\\

GeoNIRS-LP (Random Init)
& 56.12
& 57.42
& 58.07
& 58.94
\\

\midrule

RBF-SVM
& 56.05
& 57.21
& 58.06
& 58.91
\\

Stat-LDA
& 54.28
& 55.60
& 56.03
& 56.14
\\

\midrule

1D-CNN
& 52.23
& 53.51
& 55.12
& 56.51
\\

EEGNet
& 50.41
& 50.94
& 52.77
& 54.29
\\

fNIRS-T
& 55.84
& 57.24
& 59.04
& 60.36
\\

fNIRSNet
& 50.00
& 50.19
& 50.45
& 50.75
\\

fNIRS4D
& 52.29
& 54.22
& 55.71
& 57.67
\\

\midrule

MOMENT-LP
& 56.99
& 58.30
& 59.25
& 59.99
\\

MOMENT-LoRA
& 56.75
& 58.57
& 59.47
& 60.17
\\

\midrule

Jung-MAE-LP
& 57.25
& 58.55
& 60.07
& 60.68
\\

Jung-MAE-LoRA
& 57.68
& 58.54
& 60.91
& 61.73
\\

\midrule

\makecell[l]{GeoNIRS-FT vs.\\Random Init}
& \textbf{+4.72}
& \textbf{+4.04}
& \textbf{+4.34}
& \textbf{+3.30}
\\

\makecell[l]{GeoNIRS-FT vs.\\Best Non-GeoNIRS Baseline}
& \textbf{+3.02}
& \textbf{+3.20}
& \textbf{+1.78}
& \textbf{+1.81}
\\

\bottomrule
\end{tabular*}

\end{table}

\subsection{Channel-Loss Robustness}

Table~\ref{tab:channel_drop_aggregate} reports balanced accuracy averaged
equally across eight downstream tasks under fixed and random channel loss
at rates of 10\%, 25\%, and 50\%. All models are trained with complete
channel inputs; channel loss is applied only during inference.

\begin{table}[H]
\centering
\caption{Channel-loss robustness (balanced accuracy, \%).}
\label{tab:channel_drop_aggregate}

\scriptsize
\setlength{\tabcolsep}{2.5pt}
\renewcommand{\arraystretch}{1.08}

\begin{tabular*}{\linewidth}{
@{\extracolsep{\fill}}
lccccccc
@{}
}
\toprule

\textbf{Method}
& \textbf{Clean}
& \multicolumn{3}{c}{\textbf{Fixed Channel Loss}}
& \multicolumn{3}{c}{\textbf{Random Channel Loss}}
\\

\cmidrule(lr){3-5}
\cmidrule(lr){6-8}

&
& \makecell{\textbf{10\%}\\\textbf{BA ($\Delta$)}}
& \makecell{\textbf{25\%}\\\textbf{BA ($\Delta$)}}
& \makecell{\textbf{50\%}\\\textbf{BA ($\Delta$)}}
& \makecell{\textbf{10\%}\\\textbf{BA ($\Delta$)}}
& \makecell{\textbf{25\%}\\\textbf{BA ($\Delta$)}}
& \makecell{\textbf{50\%}\\\textbf{BA ($\Delta$)}}
\\
\midrule

GeoNIRS-FT
& \textbf{63.54}
& \textbf{63.21 (0.33)}
& \textbf{62.95 (0.59)}
& \textbf{62.32 (1.22)}
& \textbf{63.19 (0.35)}
& \underline{62.74 (0.80)}
& \textbf{61.84 (1.70)}
\\

GeoNIRS-LoRA
& \underline{63.42}
& \underline{63.10 (0.32)}
& \underline{62.94 (0.48)}
& \underline{62.07 (1.35)}
& \underline{63.08 (0.34)}
& 62.69 (0.73)
& \underline{61.65 (1.77)}
\\

GeoNIRS-LP
& 63.06
& 63.09 ($-0.03$)
& 62.80 (0.26)
& 62.06 (1.00)
& 62.86 (0.20)
& \textbf{62.76 (0.30)}
& 61.62 (1.44)
\\
\midrule

RBF-SVM
& 58.91
& 59.26 ($-0.35$)
& 57.45 (1.46)
& 54.35 (4.56)
& 58.25 (0.66)
& 57.50 (1.41)
& 55.24 (3.67)
\\

Stat-LDA
& 56.14
& 55.63 (0.51)
& 54.90 (1.24)
& 53.88 (2.26)
& 55.78 (0.36)
& 54.63 (1.51)
& 54.01 (2.13)
\\
\midrule

1D-CNN
& 56.51
& 56.59 ($-0.08$)
& 55.40 (1.11)
& 54.19 (2.32)
& 56.29 (0.22)
& 55.64 (0.87)
& 54.78 (1.73)
\\

EEGNet
& 54.29
& 53.94 (0.35)
& 53.28 (1.01)
& 53.29 (1.00)
& 54.04 (0.25)
& 53.38 (0.91)
& 53.39 (0.90)
\\

fNIRS-T
& 60.36
& 59.58 (0.78)
& 58.62 (1.74)
& 56.70 (3.66)
& 59.73 (0.63)
& 58.68 (1.68)
& 56.85 (3.51)
\\

fNIRSNet
& 50.76
& 50.89 ($-0.13$)
& 50.70 (0.06)
& 50.78 ($-0.02$)
& 50.75 (0.00)
& 50.72 (0.04)
& 50.65 (0.11)
\\

fNIRS4D
& 57.67
& 57.87 ($-0.20$)
& 56.74 (0.93)
& 56.50 (1.17)
& 57.73 ($-0.06$)
& 57.09 (0.58)
& 55.92 (1.75)
\\
\midrule

MOMENT-LoRA
& 60.17
& 59.32 (0.85)
& 58.04 (2.13)
& 54.18 (5.99)
& 59.14 (1.03)
& 58.05 (2.12)
& 54.52 (5.65)
\\
\midrule

Jung-MAE-LP
& 60.68
& 58.62 (2.06)
& 56.15 (4.53)
& 53.66 (7.02)
& 58.58 (2.10)
& 56.19 (4.49)
& 53.89 (6.79)
\\

Jung-MAE-LoRA
& 61.73
& 59.93 (1.80)
& 58.00 (3.73)
& 55.23 (6.50)
& 59.33 (2.40)
& 57.48 (4.25)
& 54.81 (6.92)
\\
\midrule

\makecell[l]{\textbf{GeoNIRS-FT vs.}\\\textbf{Best Non-GeoNIRS}}
& \textbf{+1.81}
& \textbf{+3.28}
& \textbf{+4.33}
& \textbf{+5.62}
& \textbf{+3.46}
& \textbf{+4.06}
& \textbf{+4.99}
\\
\bottomrule
\end{tabular*}

\vspace{2pt}
\begin{minipage}{\linewidth}
\scriptsize
\textit{Note:} Parentheses show
$\Delta_{\mathrm{drop}}=\mathrm{BA}_{\mathrm{clean}}-
\mathrm{BA}_{\mathrm{loss}}$ in percentage points.
The final row subtracts the highest non-GeoNIRS eight-task mean BA
in each column from GeoNIRS-FT.
\end{minipage}

\end{table}

\section{scaling analysis}

\subsection{Scaling the Pretraining Corpus}
\label{app:corpus_scaling}

We conduct a preliminary analysis of how downstream adaptation changes as the pretraining corpus is progressively expanded. Three nested configurations contain one, four, and five datasets, corresponding to 16.71, 26.21, and 51.96 hours of recordings, respectively. The one-dataset configuration contains \cite{medvedeva2026poststroke}; the four-dataset configuration additionally includes \cite{lee2026ds004022}, \cite{rybar2025semantic}, and \cite{sainifarron2026anesthesia}; and the five-dataset configuration further adds \cite{ajra2026cmx7mm}. All models are evaluated on the same motor-response decoding task using linear probing and full-coverage LoRA.

\begin{figure}[H]
    \centering
    \includegraphics[width=0.8\linewidth]{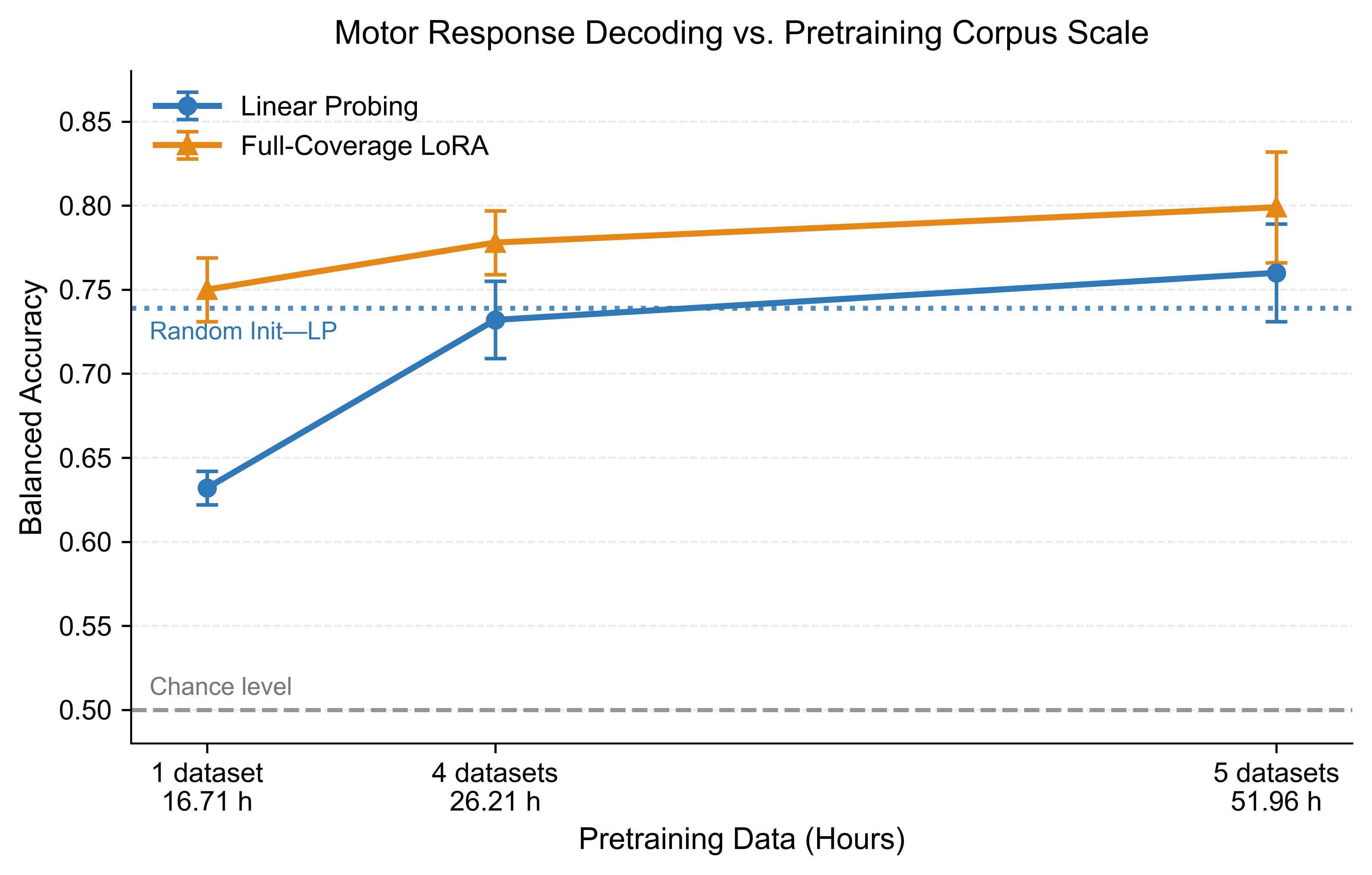}
    \caption{
    Preliminary analysis of downstream performance with increasing pretraining-corpus scale. 
    }
    \label{fig:pretraining_corpus_scaling}
\end{figure}

As shown in Figure~\ref{fig:pretraining_corpus_scaling}, performance improves consistently as the pretraining corpus expands. Linear-probing balanced accuracy increases from 63.2\% with one dataset to 73.2\% with four datasets and 76.0\% with five datasets, eventually exceeding the randomly initialized linear-probing baseline. Full-coverage LoRA similarly improves from 75.0\% to 77.8\% and 79.9\%. These results suggest that increasing the scale and diversity of the pretraining corpus improves the learned representations and their downstream adaptability. Because this pilot analysis evaluates a single downstream task and changes both corpus size and composition, it provides preliminary scaling evidence rather than a controlled scaling-law analysis.

\subsection{Scaling to GeoNIRS-large}
\label{app:model_scaling}

We further examine whether GeoNIRS benefits from an expanded model and pretraining corpus. GeoNIRS-base contains 6.5M parameters and is pretrained on the eight-dataset core corpus, whereas GeoNIRS-large contains 30.1M parameters and uses the complete 22-dataset corpus. Both variants are evaluated using the same full fine-tuning protocol. As shown in Table~\ref{tab:model_scaling_ft}, GeoNIRS-large improves balanced accuracy on six of the eight downstream tasks, including gains of 2.03 points on Action Mode, 1.20 points on Tinnitus, and 1.04 points on Semantic. Its task-averaged performance increases from 63.54\% to 63.84\%. The modest average improvement is primarily limited by a 2.79-point decrease on ICH, indicating that the expanded configuration benefits most tasks but does not yield uniform improvements across all downstream settings. Since model capacity and pretraining-corpus coverage are increased together, this experiment evaluates the overall GeoNIRS-large configuration rather than isolating a parameter-only scaling effect.

\begin{table}[H]
\centering
\caption{
Full fine-tuning performance of GeoNIRS-base and GeoNIRS-large across eight
downstream tasks. Results are balanced accuracy (\%), averaged over two
random seeds with five-fold cross-validation. $\Delta$ denotes Large minus
Base.
}
\label{tab:model_scaling_ft}

\scriptsize
\setlength{\tabcolsep}{1.5pt}
\renewcommand{\arraystretch}{1.08}

\begin{tabular*}{\columnwidth}{
@{\extracolsep{\fill}}
lcccc
@{}
}
\toprule

\textbf{Downstream Task}
& \makecell{\textbf{GeoNIRS-base}\\\textbf{(6.5M)}}
& \makecell{\textbf{GeoNIRS-large}\\\textbf{(30.1M)}}
& \boldmath$\Delta$ \textbf{(pp)}
& \makecell{\textbf{Improved?}}
\\

\midrule

ICH
& $\mathbf{80.86 \pm 8.01}$
& $78.07 \pm 11.32$
& $-2.79$
& No
\\

CI Status
& $\mathbf{69.19 \pm 10.06}$
& $68.80 \pm 11.66$
& $-0.39$
& No
\\

Tinnitus
& $57.70 \pm 8.71$
& $\mathbf{58.90 \pm 3.68}$
& $+1.20$
& \textbf{Yes}
\\

Pain Intensity
& $54.70 \pm 2.95$
& $\mathbf{54.78 \pm 2.79}$
& $+0.08$
& \textbf{Yes}
\\

Emotion
& $59.13 \pm 5.71$
& $\mathbf{59.38 \pm 4.68}$
& $+0.25$
& \textbf{Yes}
\\

Semantic
& $51.03 \pm 1.46$
& $\mathbf{52.07 \pm 2.28}$
& $+1.04$
& \textbf{Yes}
\\

Motor Recovery
& $64.75 \pm 2.27$
& $\mathbf{65.68 \pm 1.86}$
& $+0.93$
& \textbf{Yes}
\\

Action Mode
& $70.99 \pm 4.36$
& $\mathbf{73.02 \pm 5.96}$
& $+2.03$
& \textbf{Yes}
\\

\midrule

\textbf{Average}
& 63.54
& \textbf{63.84}
& $\mathbf{+0.30}$
& \textbf{6/8 tasks}
\\

\bottomrule
\end{tabular*}
\end{table}

\section{Detailed Ablation Results}
\label{app:detailed_ablations}

\subsection{Geometry perturbation after pretraining}

We first examine whether the pretrained model relies on channel geometry by perturbing its coordinate inputs after pretraining. Table~\ref{tab:pretrain_geometry_ablation_full_after} reports task-level results when channel coordinates are either replaced with zeros or shuffled across channels under full fine-tuning, LoRA, and linear probing. The pretrained checkpoint, fNIRS signals, and downstream protocols remain unchanged. Results are averaged over two random seeds, with five-fold cross-validation performed for each seed.

\begin{table}[H]
\centering
\caption{
Task-level geometry perturbation during downstream inference, measured by
balanced accuracy (\%). Zeros replaces channel coordinates with zero vectors,
whereas Shuffle permutes them across channels. Parenthesized values denote
changes from Normal in percentage points.
}
\label{tab:pretrain_geometry_ablation_full_after}

\scriptsize
\setlength{\tabcolsep}{1.2pt}
\renewcommand{\arraystretch}{1.08}

\resizebox{\linewidth}{!}{%
\begin{tabular}{
@{}
lc*{3}{ccc}cc
@{}
}
\toprule

&
&
\multicolumn{3}{c}{\textbf{Full Fine-Tuning}}
&
\multicolumn{3}{c}{\textbf{LoRA}}
&
\multicolumn{3}{c}{\textbf{Linear Probing}}
&
\multicolumn{2}{c}{\makecell{\textbf{Mean} \boldmath$\Delta$\\\textbf{(pp)}}}
\\

\cmidrule(lr){3-5}
\cmidrule(lr){6-8}
\cmidrule(lr){9-11}
\cmidrule(lr){12-13}

\textbf{Task}
& \textbf{Ch.}
& \textbf{Normal}
& \makecell{\textbf{Zeros}\\\textbf{BA ($\Delta$)}}
& \makecell{\textbf{Shuffle}\\\textbf{BA ($\Delta$)}}
& \textbf{Normal}
& \makecell{\textbf{Zeros}\\\textbf{BA ($\Delta$)}}
& \makecell{\textbf{Shuffle}\\\textbf{BA ($\Delta$)}}
& \textbf{Normal}
& \makecell{\textbf{Zeros}\\\textbf{BA ($\Delta$)}}
& \makecell{\textbf{Shuffle}\\\textbf{BA ($\Delta$)}}
& \textbf{Zeros}
& \textbf{Shuffle}
\\

\midrule

ICH
& 18
& 80.86
& 73.89 ($-6.97$)
& 74.90 ($-5.96$)
& 80.10
& 66.15 ($-13.95$)
& 65.62 ($-14.48$)
& 77.82
& 69.11 ($-8.71$)
& 63.91 ($-13.91$)
& $-9.88$
& $-11.45$
\\

CI Status
& 10
& 69.19
& 73.06 ($+3.87$)
& 71.36 ($+2.17$)
& 69.53
& 72.59 ($+3.06$)
& 70.69 ($+1.16$)
& 68.06
& 71.18 ($+3.12$)
& 70.70 ($+2.64$)
& $+3.35$
& $+1.99$
\\

Tinnitus
& 28
& 57.70
& 55.94 ($-1.76$)
& 58.82 ($+1.12$)
& 58.72
& 57.61 ($-1.11$)
& 57.35 ($-1.37$)
& 58.70
& 58.27 ($-0.43$)
& 59.51 ($+0.81$)
& $-1.10$
& $+0.19$
\\

Pain Intensity
& 23
& 54.70
& 52.61 ($-2.09$)
& 52.32 ($-2.38$)
& 54.36
& 51.60 ($-2.76$)
& 52.28 ($-2.08$)
& 56.19
& 52.64 ($-3.55$)
& 51.63 ($-4.56$)
& $-2.80$
& $-3.01$
\\

Emotion
& 66
& 59.13
& 54.99 ($-4.14$)
& 52.18 ($-6.95$)
& 59.15
& 55.37 ($-3.78$)
& 52.41 ($-6.74$)
& 58.23
& 55.26 ($-2.97$)
& 53.67 ($-4.56$)
& $-3.63$
& $-6.08$
\\

Semantic
& 6
& 51.03
& 50.84 ($-0.19$)
& 50.77 ($-0.26$)
& 51.22
& 50.87 ($-0.35$)
& 51.11 ($-0.11$)
& 50.93
& 51.83 ($+0.90$)
& 50.46 ($-0.47$)
& $+0.12$
& $-0.28$
\\

Motor Recovery
& 52
& 64.75
& 63.22 ($-1.53$)
& 61.53 ($-3.22$)
& 63.92
& 60.20 ($-3.72$)
& 60.40 ($-3.52$)
& 64.27
& 61.99 ($-2.28$)
& 60.48 ($-3.79$)
& $-2.51$
& $-3.51$
\\

Action Mode
& 41
& 70.99
& 69.66 ($-1.33$)
& 67.69 ($-3.30$)
& 70.34
& 67.98 ($-2.36$)
& 66.73 ($-3.61$)
& 70.29
& 67.14 ($-3.15$)
& 65.97 ($-4.32$)
& $-2.28$
& $-3.74$
\\

\midrule

\textbf{Average}
& --
& \textbf{63.54}
& 61.78 ($-1.76$)
& 61.20 ($-2.34$)
& \textbf{63.42}
& 60.30 ($-3.12$)
& 59.57 ($-3.85$)
& \textbf{63.06}
& 60.93 ($-2.13$)
& 59.54 ($-3.52$)
& $\mathbf{-2.34}$
& $\mathbf{-3.24}$
\\

\bottomrule
\end{tabular}
}

\end{table}

\subsection{Pretraining Component Ablations}

Table~\ref{tab:pretrain_geometry_ablation_full} reports task-level geometry
ablation results under full fine-tuning, LoRA, and linear probing.
The ablated variant removes channel coordinates from both the
channel-compression attention keys and decoder channel queries throughout
pretraining, while all other components and downstream adaptation protocols
remain unchanged. Results are averaged over two random seeds, with
five-fold cross-validation performed for each seed.

\begin{table}[H]
\centering
\caption{
Task-level ablation of channel geometry during pretraining, measured by
balanced accuracy (\%). The geometry-ablated variant removes channel
coordinates from both the channel-compression attention keys and decoder
channel queries. $\Delta$ denotes the change from GeoNIRS in percentage points.
}
\label{tab:pretrain_geometry_ablation_full}

\scriptsize
\setlength{\tabcolsep}{1.2pt}
\renewcommand{\arraystretch}{1.08}

\resizebox{\linewidth}{!}{%
\begin{tabular}{
@{}
lc
*{3}{ccc}
c
@{}
}
\toprule

&
&
\multicolumn{3}{c}{\textbf{Full Fine-Tuning}}
&
\multicolumn{3}{c}{\textbf{LoRA}}
&
\multicolumn{3}{c}{\textbf{Linear Probing}}
&
\makecell{\textbf{Mean}\\\boldmath$\Delta$}
\\

\cmidrule(lr){3-5}
\cmidrule(lr){6-8}
\cmidrule(lr){9-11}

\textbf{Task}
& \textbf{Ch.}
& \textbf{GeoNIRS}
& \makecell{\textbf{w/o}\\\textbf{Geometry}}
& \boldmath$\Delta$
& \textbf{GeoNIRS}
& \makecell{\textbf{w/o}\\\textbf{Geometry}}
& \boldmath$\Delta$
& \textbf{GeoNIRS}
& \makecell{\textbf{w/o}\\\textbf{Geometry}}
& \boldmath$\Delta$
& \textbf{(pp)}
\\

\midrule

ICH
& 18
& 80.86 & 72.44 & $-8.42$
& 80.10 & 68.09 & $-12.01$
& 77.82 & 69.26 & $-8.56$
& $-9.66$
\\

CI Status
& 10
& 69.19 & 69.31 & $+0.12$
& 69.53 & 68.84 & $-0.69$
& 68.06 & 68.66 & $+0.60$
& $+0.01$
\\

Tinnitus
& 28
& 57.70 & 57.25 & $-0.45$
& 58.72 & 57.21 & $-1.51$
& 58.70 & 56.10 & $-2.60$
& $-1.52$
\\

Pain Intensity
& 23
& 54.70 & 52.37 & $-2.33$
& 54.36 & 53.13 & $-1.23$
& 56.19 & 51.94 & $-4.25$
& $-2.60$
\\

Emotion
& 66
& 59.13 & 55.69 & $-3.44$
& 59.15 & 54.50 & $-4.65$
& 58.23 & 53.92 & $-4.31$
& $-4.13$
\\

Semantic
& 6
& 51.03 & 50.94 & $-0.09$
& 51.22 & 51.31 & $+0.09$
& 50.93 & 51.33 & $+0.40$
& $+0.13$
\\

Motor Recovery
& 52
& 64.75 & 62.00 & $-2.75$
& 63.92 & 61.22 & $-2.70$
& 64.27 & 62.00 & $-2.27$
& $-2.57$
\\

Action Mode
& 41
& 70.99 & 67.44 & $-3.55$
& 70.34 & 67.24 & $-3.10$
& 70.29 & 66.56 & $-3.73$
& $-3.46$
\\

\midrule

\textbf{Average}
& --
& \textbf{63.54} & 60.93 & $-2.61$
& \textbf{63.42} & 60.19 & $-3.23$
& \textbf{63.06} & 59.97 & $-3.09$
& $\mathbf{-2.98}$
\\

\bottomrule
\end{tabular}%
}
\end{table}

Table~\ref{tab:reconstruction_task_effect} reports the task-level changes
under each adaptation protocol when cross-chromophore reconstruction is
replaced with self-reconstruction. Cross-reconstruction improves
task-averaged balanced accuracy under all three protocols, with the
largest gains on ICH and Tinnitus. The effects vary across tasks, however,
and self-reconstruction performs better on Action Mode. Thus, the
average benefit does not imply a uniform improvement on every task.

\begin{table}[H]
\centering
\caption{
Task-level effect of replacing cross-chromophore reconstruction with
self-reconstruction. Values are changes in balanced accuracy
(Self-Reconstruction minus GeoNIRS, percentage points); negative values
favor cross-chromophore reconstruction. Results are averaged over two
seeds and five subject-level folds per seed.
}
\label{tab:reconstruction_task_effect}
\small
\setlength{\tabcolsep}{8pt}
\begin{tabular}{lrrr}
\toprule
Downstream task & Full Fine-Tuning & LoRA & Linear Probing \\
\midrule
ICH            & $-8.56$ & $-8.01$ & $-7.91$ \\
CI Status      & $+1.00$ & $-1.72$ & $+0.62$ \\
Tinnitus       & $-4.28$ & $-4.37$ & $-4.11$ \\
Pain Intensity & $+0.18$ & $-0.42$ & $-0.90$ \\
Emotion        & $-0.39$ & $-1.63$ & $-2.24$ \\
Semantic       & $+0.84$ & $-1.16$ & $+0.37$ \\
Motor Recovery & $+0.04$ & $+0.01$ & $+0.05$ \\
Action Mode    & $+1.17$ & $+1.65$ & $+1.87$ \\
\midrule
Average        & $-1.25$ & $-1.96$ & $-1.53$ \\
\bottomrule
\end{tabular}
\end{table}

\section{Discussion}

\paragraph{Why GeoNIRS works.}
The transferability of GeoNIRS may arise from three complementary factors: multi-dataset pretraining, geometry-aware channel unification, and cross-chromophore reconstruction. First, joint pretraining across datasets with different participant populations, acquisition systems, probe layouts, and experimental paradigms exposes the model to substantially broader signal variability than any individual dataset, providing a basis for learning more general fNIRS representations. The overall improvements observed when expanding the pretraining corpus are also consistent with this interpretation. Second, the geometry-aware channel compressor does more than accommodate a variable number of input channels: it uses source--detector geometry as a shared spatial reference and maps heterogeneous probe layouts into a fixed number of latent-channel tokens. Replacing channel coordinates with zeros or shuffled coordinates reduces average balanced accuracy by 2.34 and 3.24 percentage points, respectively, while removing geometry during pretraining causes a 2.98-point reduction. These results indicate that GeoNIRS meaningfully uses the correspondence between channels and their physical locations. Finally, replacing cross-chromophore reconstruction with self-reconstruction reduces average performance by 1.58 points, consistent with the motivation that the complete HbO or HbR signal provides complementary information for reconstructing its masked counterpart. GeoNIRS also outperforms the strongest non-GeoNIRS method selected separately at each label ratio and under each channel-loss condition, suggesting that the learned representations remain transferable under limited supervision and missing-channel settings rather than benefiting from comparison with a single weaker baseline.

\paragraph{Limitations.}
This study has several limitations. First, although we screened nearly 100 publicly available fNIRS datasets, only 30 provided usable signals, the required metadata, and sufficient information to recover channel geometry. The remaining datasets may still contain physiologically informative signals despite lacking reliable source--detector coordinates. Because GeoNIRS currently relies on channel geometry to establish correspondence across probe layouts, incorporating such datasets without compromising the shared spatial representation remains an important challenge for expanding the pretraining corpus. Second, several existing fNIRS pretraining methods do not provide complete implementations or downloadable pretrained weights, preventing their direct evaluation under a unified setting. Jung-inspired MAE is therefore a paper-guided implementation rather than an exact reproduction of the original method. Finally, GeoNIRS-large increases both model capacity and pretraining data, but its average improvement is modest and not consistent across all downstream tasks. The current results therefore demonstrate the potential of scaling the overall configuration but do not isolate the individual effects of model size, data volume, and corpus composition, nor do they establish a reliable scaling law for fNIRS pretraining.

\paragraph{Outlook.}
An important direction is to enable geometry-aware pretraining to benefit from both coordinate-complete and coordinate-missing fNIRS datasets. Possible approaches include geometry-optional modeling branches, estimating spatial relationships from available probe metadata, or using datasets with reliable geometry to guide representation learning on datasets without coordinates. These approaches could substantially expand the amount of data available for pretraining. Another promising direction is to integrate fNIRS with complementary modalities. EEG captures neural electrical activity at a higher temporal resolution, whereas fNIRS measures comparatively slow hemodynamic responses; joint pretraining on synchronized EEG--fNIRS recordings may therefore help model the relationship between neural activity and its vascular response. Short-separation channels and peripheral measurements such as heart rate and respiration may further help distinguish cortical hemodynamic responses from systemic physiological interference. In addition, aligning fNIRS representations with experimental stimuli, behavioral records, or textual task descriptions may support transfer across paradigms and enable few-shot or potentially zero-shot downstream applications. Future studies should also vary model capacity, data volume, and corpus composition independently to characterize how heterogeneous fNIRS pretraining scales.

\end{document}